\documentclass{article}
\usepackage{arxiv}
\usepackage{graphicx} 
\newtheorem{remark}{Remark}
\usepackage{amsmath}
\usepackage{amssymb}
\usepackage{bm}
\usepackage{mathtools}
\usepackage{xcolor}
\usepackage{subcaption}
\usepackage{comment}
\usepackage{physics}
\usepackage{subcaption}
\DeclareMathOperator*{\argmin}{arg\,min}
\usepackage[algo2e,ruled,vlined]{algorithm2e}
\DontPrintSemicolon

\title{Input Specific Neural Networks}

\author{Asghar Jadoon\\
The University of Texas at Austin\\
Austin TX, USA 
\And
D. Thomas Seidl \\
Sandia National Laboratories \\
Albuquerque NM, USA 
\And
Reese E. Jones \\
Sandia National Laboratories \\
Livermore CA, USA
\And
Jan Fuhg \\
The University of Texas at Austin\\
Austin TX, USA 
}

\date{February 2025}

\begin{document}

\maketitle

\section*{Abstract}

Neural networks have emerged as powerful tools for mapping between inputs and outputs. However, their black-box nature limits the ability to encode or impose specific structural relationships between inputs and outputs. While various studies have introduced architectures that ensure the network's output adheres to a particular form in relation to certain inputs, the majority of these approaches impose constraints on only a single set of inputs, leaving others unconstrained. This paper introduces a novel neural network architecture, termed the Input Specific Neural Network (ISNN), which extends this concept by allowing scalar-valued outputs to be subject to multiple constraints. Specifically, the ISNN can enforce convexity in some inputs, non-decreasing monotonicity combined with convexity with respect to others, and simple non-decreasing monotonicity or arbitrary relationships with additional inputs. To the best of our knowledge, this is the first work that proposes a framework that simultaneously comprehensively imposes all these constraints. The paper presents two distinct ISNN architectures, along with equations for the first and second derivatives of the output with respect to the inputs. These networks are broadly applicable.

In this work, we restrict their usage to solving problems in computational mechanics. In particular, we show how they can be effectively applied to fitting data-driven constitutive models. We remark, that due to their increased ability to implicitly model constraints, we can show that ISNNs require fewer inputs than existing input convex neural networks when modeling polyconvex hyperelastic functions. 
We then embed our trained data-driven constitutive laws into a finite element solver where significant time savings can be achieved by using explicit manual differentiation using the derived equations as opposed to automatic differentiation. Manual differentiation also enables seamless employment of trained ISNNs in commercial solvers where automatic differentiation may not be possible. We also show how ISNNs can be used to learn structural relationships between inputs and outputs via a binary gating mechanism. Particularly, ISNNs are employed to model an anisotropic free energy potential to get the homogenized macroscopic response in a decoupled multiscale setting, where the network learns whether or not the potential should be modeled as polyconvex, and retains only the relevant layers while using the minimum number of inputs.

\section{Introduction}

Neural networks (NNs) have proven to be reliable tools for mapping functions between inputs and outputs, and have found applications in a wide array of disciplines including, but not limited to, engineering, finance, and social sciences \cite{kaveh2024applications, fuhg2024review, gogas2021machine, grimmer2021machine}. However, a common critique of NNs is their black-box nature \cite{flaschel2021unsupervised, di2021machine}, i.e., the ability to obtain simple functional expressions to link the inputs and outputs is lost due to overparameterization. Consequently, it is generally also not possible to prove structural relationships, e.g. convexity, between the inputs and the outputs. This is a major issue in the realm of scientific machine learning since structural relationships are often known and could be leveraged. These generally stem from physical principles or assumptions involved in deriving mathematical tools to solve physical systems \cite{linden2023neural, jadoon2025automated, meyer2023thermodynamically, Fuhg2024-fz}.

 One option to incorporate the constraints is by employing a form of \textit{informed} neural network \cite{xu2023practical, arzani2021uncovering}. These rely on modifying the loss term such that the violation of the preestablished structural relationships is penalized. In a sense, these networks only softly enforce the constraints \cite{Fuhg2023-eb} and they might be violated especially outside of the domain spanned by the training data.

 A more effective, albeit involved, method relies on constructing neural network architectures that, by construction, enforce these structural relationships. By doing so, the hope is that not only can the training data be reduced, but the network is also expected to have better generalization to out-of-distribution data. Some examples of these networks include \textit{input convex} or \textit{partially input convex} neural networks \cite{amos2017input} and \textit{monotone} or \textit{partially monotone} networks \cite{daniels2010monotone}. As the names suggest, these neural network architectures enforce convexity or monotonicity with respect to some inputs while allowing arbitrary relationships to other inputs.

However, to the best of the authors' knowledge, the majority of such network architectures only impose a single constraint on a set of inputs, leaving others unconstrained. In this work, we present a novel neural network termed \textit{input specific neural networks} (ISNNs) that is able to impose multiple, different constraints on different sets of inputs for a scalar-valued output. More specifically, ISNNs can enforce convexity to some inputs, monotonicity combined with convexity with respect to others, and simple monotonicity or arbitrary relationships with additional inputs. This is the first work to impose multiple constraints in this comprehensive manner simultaneously. Two architectures are presented for ISNNs with varying complexity along with their analytical derivatives. Along with these architectures, constraints that must be enforced are presented in order to preserve the required structural forms of the output with respect to the given inputs.

This paper is organized as follows. The proposed ISNN architectures are presented in section 2. These architectures are then put to test in section 3 where the results are compared with existing neural network formulations. Particularly, we start off with two toy problems. We then turn to computational mechanics and solve an inverse problem in isotropic hyperelasticity where we show that the ISNNs outperform partially input convex NNs. We highlight that we are able to reduce the number of required invariant inputs for the forward problem under polyconvex assumptions. A comparative study of the computational cost of manual and automatic derivatives is also presented. Furthermore, section 3 also introduces an ISNN formulation that can learn structural relationships between the inputs and the output, in case these are not known \textit{a priori}. The paper ends with a brief discussion of the results and the conclusions in section 4.

\section{Neural Network Architecture}
\begin{figure}
    \centering
    \includegraphics[scale = 0.35]{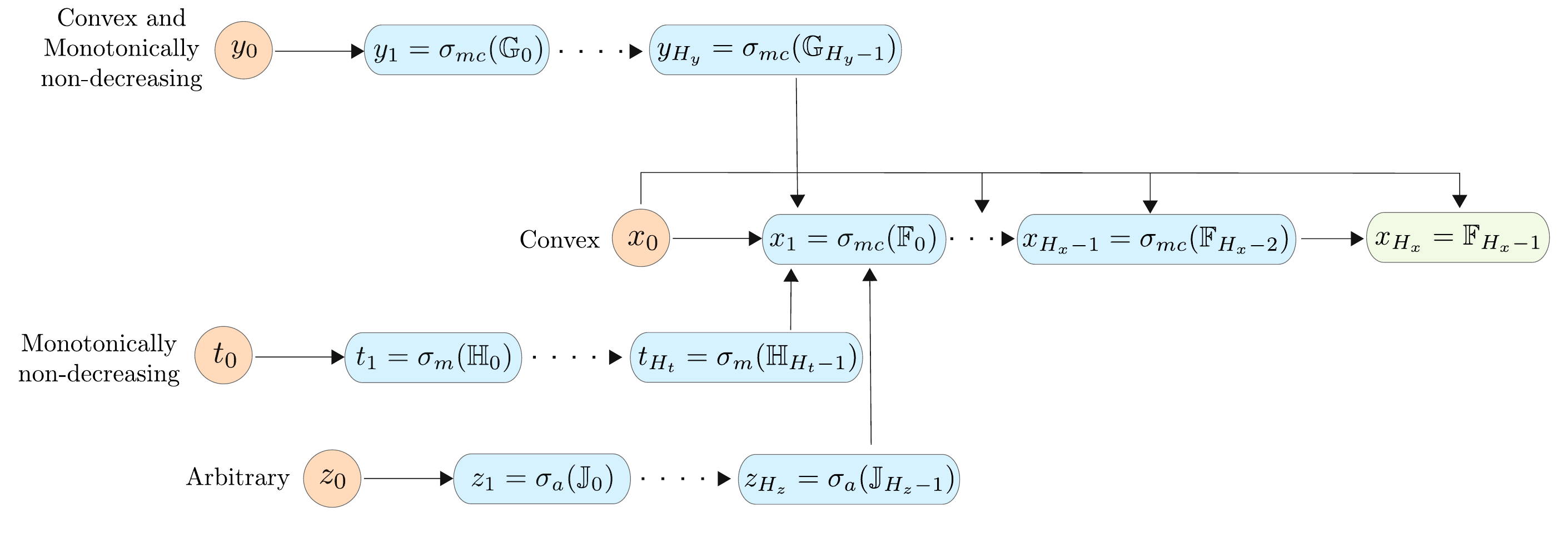}
    \caption{ISNN-1 schematic with the terms inside the activation functions corresponding to those from Eq. \eqref{yHtype1_1} to Eq. \eqref{xHtype1_1}.}
    \label{fig:ISNNtype1}
\end{figure}

In this section, we present two architectures for ISNNs. These architectures vary in complexity and the choice of architecture can be problem specific. Figures \ref{fig:ISNNtype1} and \ref{fig:ISNNtype2} show an illustration of each network whereas the following sections will contain a more mathematical treatment of them. The analytical derivatives (first and second order) for each network are presented in Appendix \ref{AppA}, along with the constraints to attain desired structural forms between inputs and output.

In the following $\mathbf{W}$ and $\mathbf{b}$ are trainable weight matrices and bias vectors, respectively which form learnable affine transforms such as $\mathbb{F}$ and $\mathbb{G}$. The symbol $\sigma$ denotes an activation function with the subscripts highlighting different constraints imposed on them as will be explained subsequently. In both architectures, the network is convex with respect to inputs $\mathbf{x}_0$, convex and monotonically non-decreasing with respect to $\mathbf{y}_0$, only monotonically non-decreasing with $\mathbf{t}_0$ and arbitrary with respect to $\mathbf{z}_0$.

\subsection*{ISNN-1: Type 1 input specific neural network}

The network takes as input $\mathbf{x}_0$, $\mathbf{y}_0$, $\mathbf{t}_0$, $\mathbf{z}_0$ and gives the output $P(\mathbf{x}_0$, $\mathbf{y}_0$, $\mathbf{t}_0$, $\mathbf{z}_0) \coloneqq x_{H_x} \in \mathbb{R}$. 

\begin{equation}
\mathbf{y}_{h+1} = \sigma_{mc} \underbrace{\left(\mathbf{y}_h  \mathbf{W}_h^{[yy]^T} + \mathbf{b}_h^{[y]} \right)}_{\mathbb{G}_h} , \quad h = 0, \ldots, H_y - 1
\label{yHtype1_1}
\end{equation}

\begin{equation}
\mathbf{z}_{h+1} = \sigma_a \underbrace{\left( \mathbf{z}_h \mathbf{W}_h^{[zz]^T} + \mathbf{b}_h^{[z]} \right)}_{\mathbb{J}_h} , \quad h = 0, \ldots, H_z - 1
\label{zHtype1_1}
\end{equation}

\begin{equation}
\mathbf{t}_{h+1} = \sigma_m \underbrace{\left(\mathbf{t}_h  \mathbf{W}_h^{[tt]^T} + \mathbf{b}_h^{[t]} \right)}_{\mathbb{H}_h} , \quad h = 0, \ldots, H_t - 1
\label{tHtype1_1}
\end{equation}

\begin{equation}
\mathbf{x}_1 = \sigma_{mc} \underbrace{\left( \mathbf{x}_0 \mathbf{W}_0^{[xx]^T} + \mathbf{b}_0^{[x]} + \mathbf{y}_{H_y}  \mathbf{W}^{[xy]^T} + \mathbf{z}_{H_z} \mathbf{W}^{[xz]^T}  + \mathbf{t}_{H_t} \mathbf{W}^{[xt]^T}  \right)}_{\mathbb{F}_0}
\label{x0type1_1}
\end{equation}

\begin{equation}
\mathbf{x}_{h+1} = \sigma_{mc} \underbrace{\left( \mathbf{x}_h \mathbf{W}_h^{[xx]^T} + \mathbf{b}_h^{[x]} \right)}_{\mathbb{F}_h} , \quad h = 1, \ldots, H_x - 1
\label{xHtype1_1}
\end{equation}

\begin{itemize}

\item Convex with respect to $\mathbf{x}_0$ given $\mathbf{W}_h^{[xx]}$ is non-negative for $h = 1, \ldots, H_x - 1$ and $\sigma_{mc}$ is a convex, monotonically non-decreasing function.
\item Convex and monotonically non-decreasing with respect to $\mathbf{y}_0$ given $\mathbf{W}_h^{[yy]}$ and $\mathbf{W}_h^{[xy]}$ are non-negative and $\sigma_{mc}$ is a convex, monotonically non-decreasing function.
\item Arbitrary with respect to $\mathbf{z}_0$. For this case, $\sigma_a$ can take any form and there are no restrictions on $\mathbf{W}_h^{[zz]}$ and $\mathbf{W}_h^{[xz]}$.
\item Monotonically increasing with respect to $\mathbf{t}_0$, but not necessarily convex. Both $\mathbf{W}_h^{[tt]}$ and $\mathbf{W}_h^{[xt]}$ have to be non-negative whereas $\sigma_m$ needs only to be a monotonically non-decreasing function.

\end{itemize}







\subsection*{ISNN-2: Type 2 input specific neural network}

The network takes as input $\mathbf{x}_0$, $\mathbf{y}_0$, $\mathbf{t}_0$, $\mathbf{z}_0$ and gives the output $P(\mathbf{x}_0$, $\mathbf{y}_0$, $\mathbf{t}_0$, $\mathbf{z}_0) \coloneqq x_{H} \in \mathbb{R} $.

\begin{equation}
\mathbf{y}_{h+1} = \sigma_{mc}\underbrace{ \left(  \mathbf{y}_h \mathbf{W}_h^{[yy]^T} + \mathbf{b}_h^{[y]} \right)}_{\mathbb{G}_h} , \quad h = 0, \ldots, H-2
\label{y_eq_2}
\end{equation}

\begin{equation}
\mathbf{z}_{h+1} = \sigma_a \underbrace{ \left(\mathbf{z}_h  \mathbf{W}_h^{[zz]^T} + \mathbf{b}_h^{[z]} \right)}_{\mathbb{J}_h} , \quad h = 0, \ldots, H-2
\label{z_eq_2}
\end{equation}

\begin{equation}
\mathbf{t}_{h+1} = \sigma_m \underbrace{ \left( \mathbf{t}_h \mathbf{W}_h^{[tt]^T} + \mathbf{b}_h^{[t]} \right) }_{\mathbb{H}_h}, \quad h = 0, \ldots, H-2
\label{t_eq_2}
\end{equation}

\begin{equation}
\mathbf{x}_{1} = \sigma_{mc} \underbrace{ \left( \mathbf{x}_0 \mathbf{W}_0^{[xx]^T} + \mathbf{y}_0 \mathbf{W}_0^{[xy]^T}  + \mathbf{z}_0 \mathbf{W}_0^{[xz]^T} + \mathbf{t}_0 \mathbf{W}_0^{[xt]^T} + \mathbf{b}_0^{[x]} \right)}_{\mathbb{F}_0}
\label{x_eq_2_0}
\end{equation}

\begin{equation}
\mathbf{x}_{h+1} = \sigma_{mc} \underbrace{ \left( \mathbf{x}_h \mathbf{W}_h^{[xx]^T} +  \mathbf{x}_0 \mathbf{W}_h^{[xx_{0}]^T} + \mathbf{y}_h \mathbf{W}_h^{[xy]^T}  + \mathbf{z}_h \mathbf{W}_h^{[xz]^T} + \mathbf{t}_h \mathbf{W}_h^{[xt]^T} + \mathbf{b}_h^{[x]} \right)}_{\mathbb{F}_h} , \quad h = 1, \ldots, H - 1
\label{x_eq_2}
\end{equation}

\begin{figure}
    \centering
    \includegraphics[scale = 0.35]{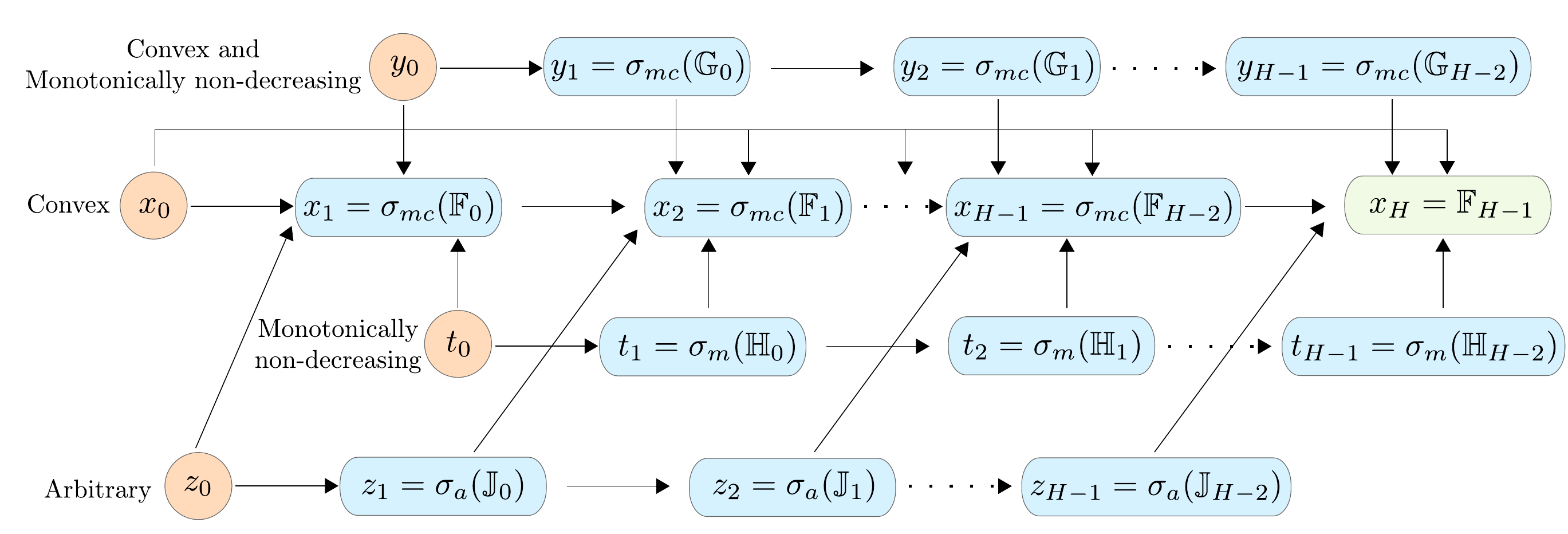}
    \caption{ISNN-2 schematic with the terms inside the activation functions corresponding to those from Eq. \eqref{y_eq_2} to Eq. \eqref{x_eq_2}.}
    \label{fig:ISNNtype2}
\end{figure}

\begin{itemize}

\item Convex with respect to $\mathbf{x}_0$ given $\mathbf{W}_h^{[xx]}$ is non-negative for $h = 1, \ldots, H$ and $\sigma_{mc}$ is a convex, monotonically non-decreasing function whereas $\mathbf{W}_h^{[xx0]}$ can take any value.
\item Convex and monotonically non-decreasing with respect to $\mathbf{y}_0$ given $\mathbf{W}_h^{[yy]}$ and $\mathbf{W}_h^{[xy]}$ are non-negative for $h = 0, \ldots, H$ and $\sigma_{mc}$ is a convex, monotonically non-decreasing function.
\item Arbitrary with respect to $\mathbf{z}_0$. As before, $\sigma_a$ can take any form and there are no restrictions on $\mathbf{W}_h^{[zz]}$ and $\mathbf{W}_h^{[xz]}$.
\item Monotonically increasing with respect to $\mathbf{t}_0$, but not necessarily convex. Both $\mathbf{W}_h^{[tt]}$ and $\mathbf{W}_h^{[xt]}$ have to be non-negative for $h = 0, \ldots, H$ whereas $\sigma_m$ needs only to be a monotonically non-decreasing function.
\item All the biases can be negative or positive and do not affect any of the constraints.
\end{itemize}

\begin{remark}
Note that the output from the last layer of the neural network is generally not placed inside an activation function. Here, the activation function is used with the output to avoid having to write the last layer separately. However, even without the activation function in the output layer, the constraints are not violated. The networks illustrated in Figures \ref{fig:ISNNtype1} and \ref{fig:ISNNtype2} do not have the activation function around the last layer.
\end{remark}

\begin{remark}
The so-called \textit{pass-through layers}/\textit{skip connections}  generally help with the representative power of the network. In the architecture presented, they are only used for the inputs $\mathbf{x}_0$. However, they can also be used for other inputs if needed, with due consideration towards the constraints.
\end{remark}

\begin{remark}
Generally, a problem will not require all the constraints enforced in the neural networks presented here. In case a convexity constraint with respect to some inputs is present, the network architecture can easily be modified to remove other inputs/constraints e.g. by setting all the weights and biases associated with those inputs to zero. However, the modification of the neural network might be more involved if there are no inputs requiring convexity. For example, \cite{klein2025neuralnetworksmeethyperelasticity} used a monotonic network to model isotropic hyperelastic potentials with relaxed ellipticity conditions. Nevertheless, similar techniques as shown in the architecture presented here can be applied to pass through the information between the network branches associated with different inputs.
\end{remark}

\section{Applications}
In this section, we present applications of ISNNs. Specifically, we start off by comparing the extrapolation capabilities of ISNNs with Feed-Forward Neural Networks (FFNNs) for two toy problems. Then, we employ ISNNs to solve inverse problems in finite strain hyperelasticity, comparing the results with the current state-of-the-art methodologies for such problems. Next, we embed a trained ISNN into a finite element solver to analyze 3D Cook's membrane under specified boundary conditions. We furthermore compare the required computing time for obtaining first and second derivatives of the network through automatic and manual differentiation. We report significant time savings by evaluating the derivatives manually. As a final application, we use ISNNs to actually learn the structural relationships between the inputs and outputs. Specifically, ISNNs are used to assess whether a given stress-strain dataset can be modeled through a polyconvex potential or not. All the following models are implemented in PyTorch \cite{NEURIPS20199015} and employ the Adam optimizer \cite{kingma2014adam} for training. As activation functions, in the following, we use:

\begin{equation}
  \sigma_{mc}(x) = \log(1+\exp (x)) \ , \quad \sigma_m(x) = \sigma_a(x) = \frac{1}{1+ \exp (-x)}  \ ,
\end{equation}

which fulfill the constraint requirements.


\subsection{Toy problems}

\begin{figure}
    \begin{subfigure}{0.5\linewidth}
   \centering
        \includegraphics[scale=0.35]{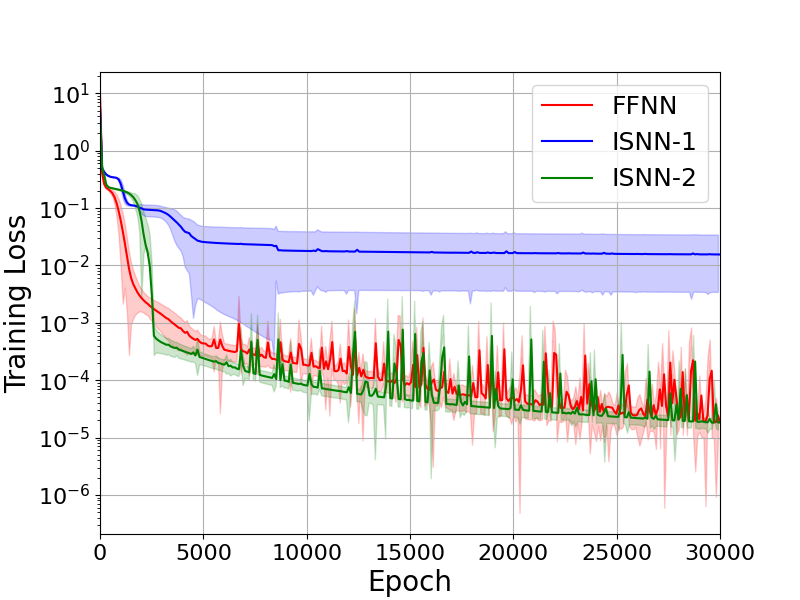}
    \caption{Training Loss}\label{fig:fun_train_loss}
    \end{subfigure}
        \begin{subfigure}{0.5\linewidth}
      \centering
        \includegraphics[scale=0.35]{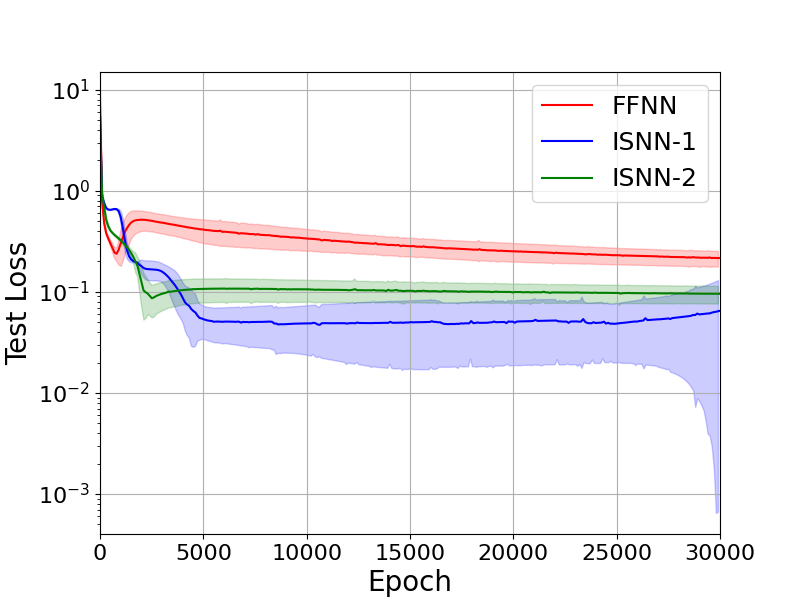}
    \caption{Test loss}\label{fig:fun_test_loss}
    \end{subfigure}
    \caption{(a) Training loss and the (b) test loss for both ISNN architectures and the FFNN for the dataset generated using Eq. \eqref{eq_f}. For each architecture, the solid line shows the mean loss over 10 different initializations whereas the shaded region shows the standard deviation.}
    \label{loss_fun}
\end{figure}
\begin{figure}
        \begin{subfigure}{0.5\linewidth}
        \centering
        \includegraphics[scale=0.35]{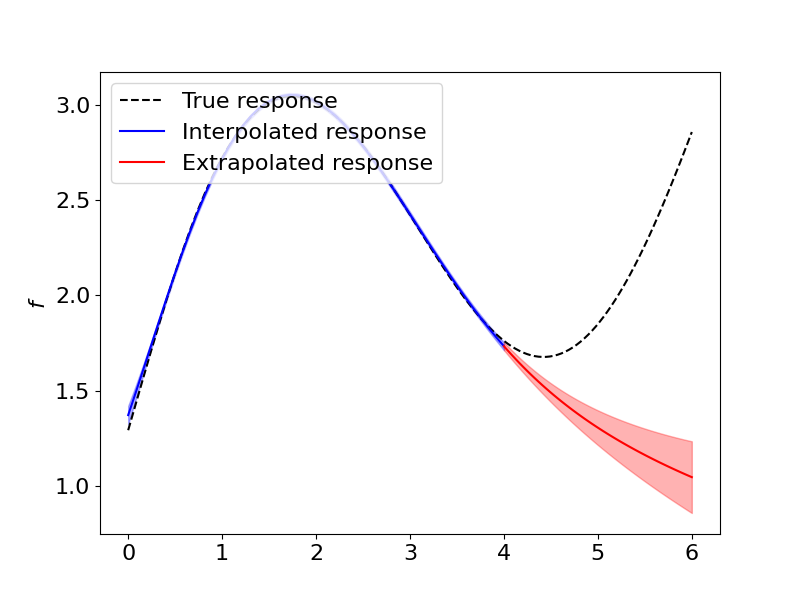}
    \caption{FFNN}\label{fig:Response_FFNN_fun}
    \end{subfigure}
        \begin{subfigure}{0.5\linewidth}
        \centering
        \includegraphics[scale=0.35]{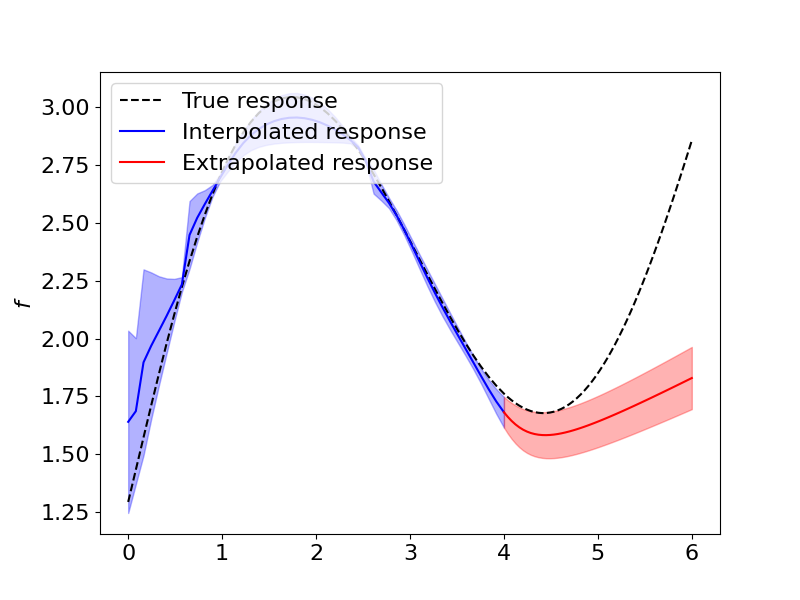}
    \caption{ISNN-1}\label{fig:Response_ISNN_1_fun}
    \end{subfigure}
    \begin{center}
            \begin{subfigure}{0.5\linewidth}
        \centering
        \includegraphics[scale=0.35]{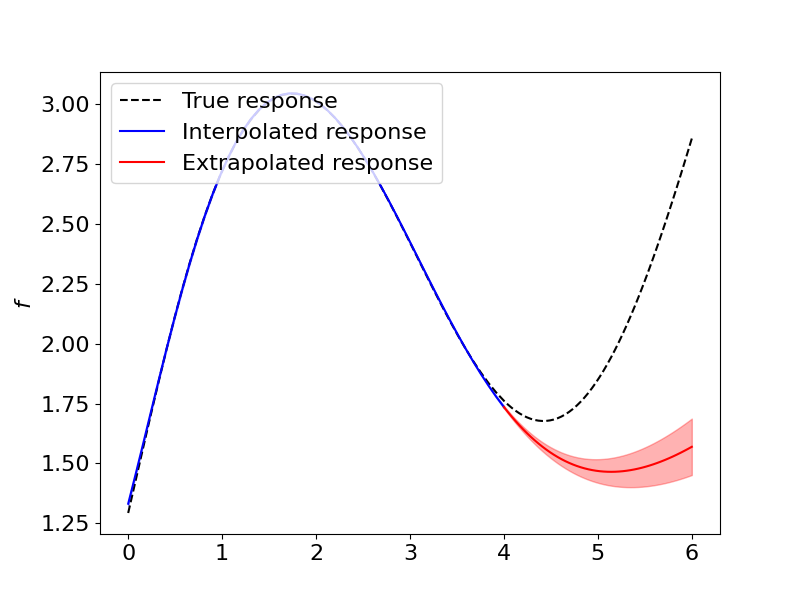}
    \caption{ISNN-2}\label{fig:Response_ISNN_2_fun}
    \end{subfigure}
    \end{center}
    \caption{Each model's predictive behavior for data seen during training, denoted with \textit{Interpolated response} and on unseen data denoted with \textit{Extrapolated response} for the dataset generated using Eq. \eqref{eq_f}. For each architecture, the solid line shows the mean loss over 10 different initializations whereas the shaded region shows the standard deviation.}
    \label{funResults}
\end{figure}

Consider the following additively split function:
\begin{equation}
    f = \exp(-0.5 x) + \log(1 + \exp(0.4 y)) + \tanh(t) + \sin(z)  - 0.4 \label{eq_f} .
\end{equation}
Clearly this function is convex in $x$, convex and monotonically increasing w.r.t. $y$, monotonically increasing w.r.t. $t$ and arbitrary w.r.t. $z$. We employ ISNNs to fit the response and evaluate the extrapolation behavior. The results from both architectures are compared with those from an unconstrained FFNN. We generate the training dataset by evaluating Eq. \eqref{eq_f} after sampling 500 times in the domain $x,y,t,z \in \{ 0.0, 4.0 \}$ using Latin Hypercube Sampling (LHS) \cite{stein1987large}. The test dataset contains 5000 samples and is generated for $x,y,t,z \in \{ 0.0, 6.0 \}$ again using LHS. We trained an FFNN and both ISNN architectures on this data for 10 different initializations of network weights and biases. The number of trainable parameters for FFNN was 2041 with 2 hidden layers with 30 neurons each whereas ISNN-1 had 1600 trainable parameters with two hidden layers for each input with 10 neurons in each layer. ISNN-2 contained 1 hidden layer for each input with 15 neurons each adding to a total of 1877 trainable parameters. The training loss is shown in Figure \ref{fig:fun_train_loss}, while Figure \ref{fig:fun_test_loss} compares the performance of each model on the test data. We see that the training loss is comparable for FFNN and ISNN-2 while ISNN-1 performs worse. However, its test loss is better compared to the other two architectures. We can see that both ISNN architectures are able to outperform FFNN on test data. However, its lower error on the training data might make ISNN-2 the preferred choice. This is also evident from Figure \ref{funResults} where we evaluate each trained model for $x=y=t=z \in \{ 0.0, 6.0\}$. We can see that FFNN is worse in extrapolation. The ISNN architectures seem to extrapolate better, potentially, due to their embedded constraints.

\begin{figure}
    \begin{subfigure}{0.5\linewidth}
        \includegraphics[scale=0.35]{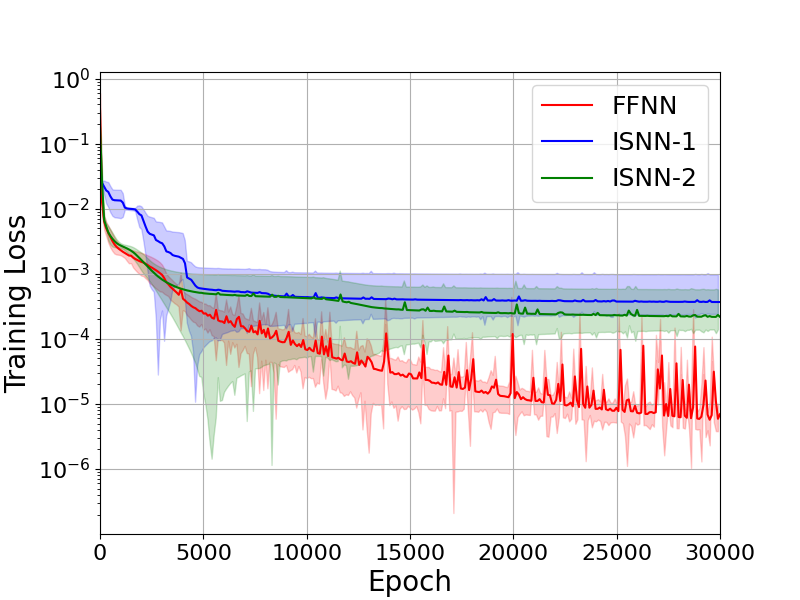}
    \caption{}
    \end{subfigure}
        \begin{subfigure}{0.5\linewidth}
        \includegraphics[scale=0.35]{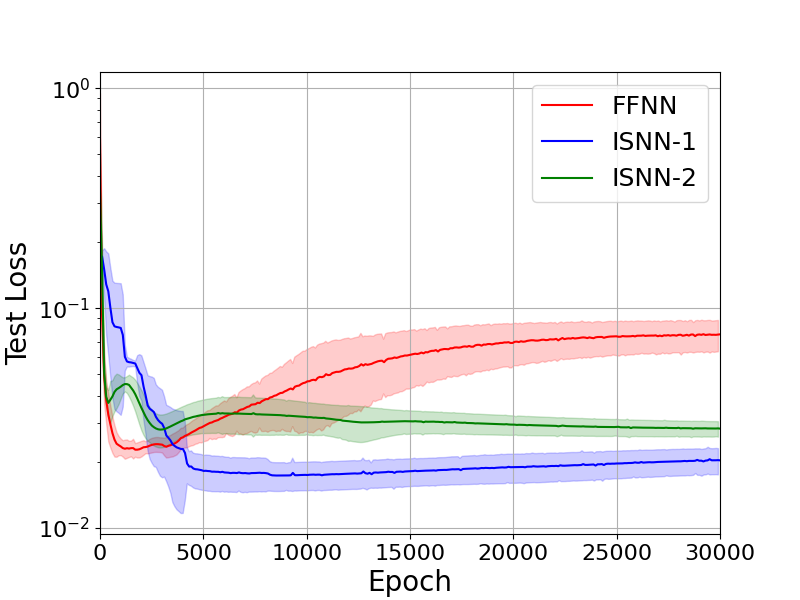}
    \caption{}
    \end{subfigure}
    \caption{(a) Training and (b) test losses for both ISNN architectures and the FFNN for the dataset generated using Eq. \eqref{eq_g}. For each architecture, the solid line shows the mean loss over 10 different initializations whereas the shaded region shows the standard deviation.}
    \label{new_func_loss}
\end{figure}

\begin{figure}
        \begin{subfigure}{0.5\linewidth}
        \centering
        \includegraphics[scale=0.35]{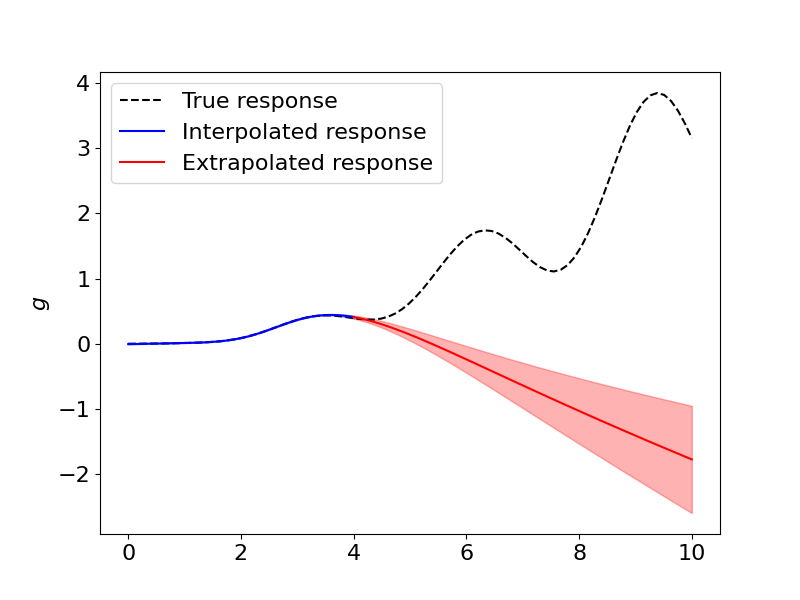}
    \caption{FFNN}\label{fig:Response_FFNN_NEWfun}
    \end{subfigure}
        \begin{subfigure}{0.5\linewidth}
        \centering
        \includegraphics[scale=0.35]{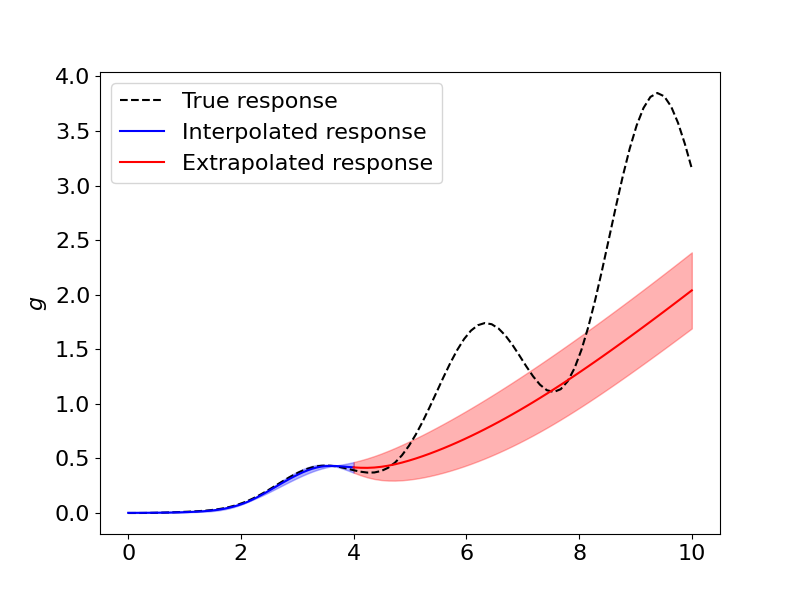}
    \caption{ISNN-1}\label{fig:Response_ISNN_1_NEWfun}
    \end{subfigure}
    \begin{center}
            \begin{subfigure}{0.5\linewidth}
            \centering
        \includegraphics[scale=0.35]{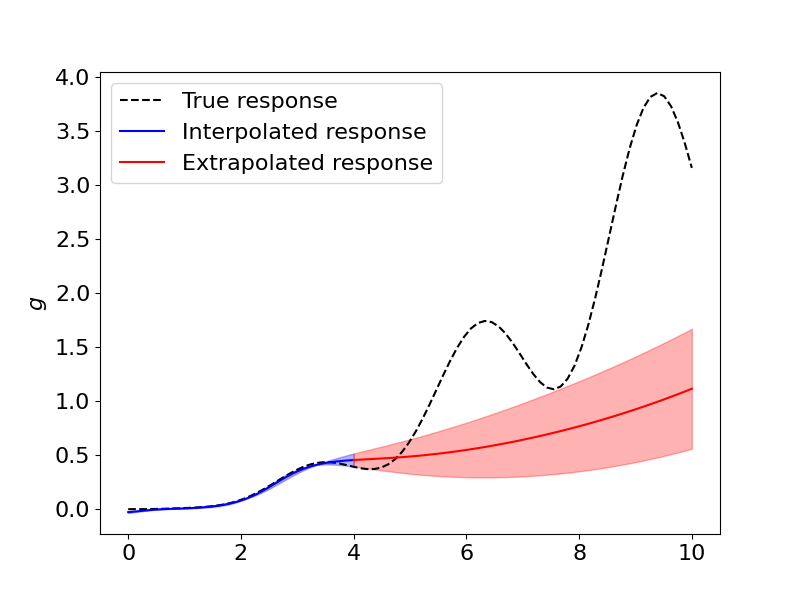}
    \caption{ISNN-2}\label{fig:Response_ISNN_2_NEWfun}
    \end{subfigure}
    \end{center}
    \caption{Each model's predictive behavior for data seen during training, denoted with \textit{Interpolated response} and on unseen data denoted with \textit{Extrapolated response} for the dataset generated using Eq. \eqref{eq_g}. For each architecture, the solid line shows the mean loss over 10 different initializations whereas the shaded region shows the standard deviation.}
    \label{new_funcResults}
\end{figure}

Next, consider the following multiplicatively split function:
\begin{equation}
    g(x,y,z,t) = f_x \cdot f_y \cdot f_z \cdot f_t \label{eq_g} \ ,
\end{equation}
where,
\begin{equation}\label{eq:f_z}
    \begin{aligned}
    f_x &= \exp(-0.3x), \quad 
    &&f_y = (0.15y)^2 \\
    f_t &= \tanh(0.3t), \quad 
    &&f_z = 0.2\sin(0.5z+2)+0.5  \ .
\end{aligned}
\end{equation}
The function $g$ is convex in $x$, convex and monotonically increasing in $y$, monotonically increasing in $t$, and arbitrary in $z$. The training and test datasets are generated similarly as before, with the only difference being the range for test data which is now set to $x,y,t,z \in \{ 0.0, 10.0 \}$. We again train each network with 10 different initializations of weights and biases. The training loss for all three architectures along with the test loss is shown in Figure \ref{new_func_loss}. We see that both ISNN architectures perform comparably during training whereas FFNN seems to overfit the training data. This is then evident in the test loss where ISNN architectures outperform FFNN. The responses for each model are shown in Figure \ref{new_funcResults} for the function $g$ evaluated at $x=y=t=z \in \{ 0.0, 10.0 \}$. Due to the complexity of the function, it appears difficult for all networks to obtain accurate fits in the extrapolation regime. However, we obverse that the ISNNs are still able to capture the general trend of the function due to their constrained nature whereas the FFNN appears unable to follow the functional response.

\subsection{Inverse problems in hyperelasticity} \label{InverseIsotropy}
Next, we employ ISNNs to solve inverse problems in isotropic hyperelasticity. We compare their performance against partially input convex neural networks (pICNNs) which have previously been employed \cite{klein2023parametrized, jadoon2024inverse} to solve such problems. We will provide a brief overview. Interested readers are referred to the aforementioned references for a more detailed formulation. The mechanical response of hyperelastic materials is generally defined by a strain energy potential $\Psi$ which depends on the deformation gradient $\mathbf{F}$ and some material or design parameters $\mathcal{D}$. The ellipticity of this potential ensures material stability \cite{Zee1983-ml}. Since ensuring ellipticity is not straightforward, we ensure polyconvexity of the potential as that guarantees sequential weak lower semicontinuity. Although it is a stronger condition, a polyconvex potential, along with coercivity ensures the existence of a minimizer in elastostatics \cite{ball1976convexity}. 

A free energy potential is polyconvex if and only if there exists a function $\mathcal{P}$ such that
\begin{equation}
  \Psi(\mathbf{F}, \mathcal{D}) = \mathcal{P}(\mathbf{F}, \text{Cof} \,\mathbf{F}, \det \mathbf{F}, \mathcal{D}) ,
\end{equation}
where $\mathcal{P}$ is convex with regard to the minors of $\mathbf{F}$, i.e., $\mathbf{F}$, \text{Cof} $\mathbf{F}$, and $\det \mathbf{F}$. From the strain energy function, the second Piola-Kirchhoff stress can be obtained with $\mathbf{S}  = 2 \partial_\mathbf{C} \Psi$.

In the case of isotropy, the polyconvex potential can be formulated in terms of the invariants of the Cauchy-Green deformation tensor ${\mathbf{C}} = \mathbf{F}^T \mathbf{F}$. This also allows us to adhere to material frame indifference and symmetry constraints.
Interested readers are referred to \cite{linden2023neural} for an overview of all physical and mechanistic constraints of the polyconvex potential and how to enforce them in a neural network setting. 

We design the NN-based potential $\Psi_{NN}$ by specifying the invariants $\mathcal{I} = \{ \mathcal{I}_1 = \text{tr}\mathbf{C} , \mathcal{I}_2 = \text{tr}\text{Cof}\mathbf{C} , \mathcal{I}_3 = J = \det \mathbf{F} \}$ as its input. The complete potential then takes the form:
\begin{equation}
    \Psi = \Psi_{NN}(\mathcal{I}_1, \mathcal{I}_2, \mathcal{I}_3) - \Psi_{NN}(3,3,1) - \mathfrak{o}(\mathcal{I}_3 - 1) \ , .
\end{equation}
The second term ensures the potential is zero at the undeformed configuration and
\begin{equation}
    \mathfrak{o} = 2 \left (\partial_{\bar{\mathcal{I}}_1}  \Psi + 2\partial_{\bar{\mathcal{I}}_2} \Psi + \frac{1}{2}\partial_{\bar{\mathcal{I}}_3}  \right) ,
\end{equation}
with $\bar{\mathcal{I}}_1$, $\bar{\mathcal{I}}_2$ and $\bar{\mathcal{I}}_3$ evaluated at $\mathbf{C}=\mathbf{I}$ ensures a stress-free condition at the undeformed configuration. Since this term is a linear function of $\mathcal{I}_3=J$, we preserve polyconvexity of the potential. The growth condition can also be added to the potential \cite{linden2023neural}, however, it is rather theoretical in nature \cite{klein2022polyconvex} since we never practically experience cases where $\mathcal{I}_3$ would go to zero or infinity.

To ensure polyconvexity, we need to be convex and monotonically increasing in $\mathcal{I}_1$ and $\mathcal{I}_2$ \cite{Boyd_Vandenberghe_2011} and \textit{only} convex in $\mathcal{I}_3$. With the existing pICNN architectures, it is not possible to simply enforce these two constraints. 
In particular, as reported in \cite{linden2023neural}, existing architectures can only be either convex or monotonically non-decreasing in a subset of its inputs (and free in the remaining ones).
Therefore, currently, in order to conform with the polyconvexity constraints, all the invariants ($I_{1}, I_{2}$ \textbf{and} $J$) are modeled as convex, monotonically non-decreasing in all existing works of the literature, see e.g. \cite{stollberg2025multiscale,jones2025attention}. However, due to this overconstraint, these models are unable to represent negative stresses \cite{linden2023neural, klein2022polyconvex} and therefore add an additional invariant ($-2J$) to the input set. This means that instead of three inputs ($I_{1}, I_{2}, J$) the neural networks require four inputs ($I_{1}, I_{2}, J, -2J$) leading to larger network sizes and less intuitive modeling from a practitioner's perspective.
One of the big advantages of the presented ISNNs is that there is no such requirement. 
Furthermore, the network can be arbitrary with respect to the design parameters $\mathcal{D}$, or if we know the structural relationship between (some of) the design parameters and the free energy $\textit{a priori}$, we can also pass it to the corresponding branch of the ISNN. 

For our inverse problem, we intend to find the design parameters that correspond to a particular stress response. In particular, assume we aim for target stresses $\mathbf{S}^{\star}$ at particular strains $\mathcal{I}^\star$. Given a training dataset $ \lbrace (\mathcal{I}^{\dagger}, \mathcal{D}^{\dagger}), \mathbf{S}^{\dagger}  \rbrace$, we therefore try to identify the design parameter $\mathcal{D}^{\star}$ that correspond to the relationship $\mathcal{I}^\star - \mathbf{S}^{\star}$. We can solve this problem by using physics-augmented neural networks as recently proposed in \cite{jadoon2024inverse}.

We formalize this through the following optimization problem:
\begin{equation}
\begin{aligned}
       \argmin_{\mathcal{D}^{\star}} \, \, &\| \mathbf{S}^{\star} - \hat{\mathbf{S}}^{\star}(\mathcal{I}^\star , \mathcal{D}^{\star}; \bm{\theta}) \|_2^2 \\
       \text{s.t.} \argmin_{\bm{\theta}} \, \, &\| \mathbf{S}^{\dagger} - \hat{\mathbf{S}}^{\dagger}(\mathcal{I}^{\dagger}, \mathcal{D}^{\dagger}; \bm{\theta}) \|_2^2 ,
\end{aligned}
\end{equation}
where $\bm{\theta}$ are the trainable neural network parameters. The hat ($\hat{\bullet}$) denotes predicted stresses obtained from NN-based potentials. We solve the optimization problem sequentially by breaking the problem down into two subproblems: the forward problem and the inverse problem. The forward problem involves training the neural networks to the dataset of the form $\lbrace (\mathbf{C}^\dagger,\mathcal{D}^\dagger), \mathbf{S}^\dagger\rbrace$. In this work, we generate the data using the Blatz and Ko model \cite{blatz1963application, holzapfel2002nonlinear} which describes a coupled representation of the isochoric and volumetric parts of a free energy potential
\begin{equation}
    \Psi(\mathcal{I}_1, \mathcal{I}_2, J, \mu, \beta) =  \frac{\mu}{4} \left[(\mathcal{I}_1 - 3) + \frac{1}{\beta} (J^{-2\beta}-1) \right] +\frac{\mu}{4} \left[\left(\frac{\mathcal{I}_2}{J^2} - 3 \right) + \frac{1}{\beta}(J^{2\beta} -1) \right] ,
    \label{psi_blatzko}
\end{equation}
where $\mu$ is the shear modulus and $\beta = \frac{\nu}{1-2\nu}$ with $\nu$ being the Poisson's ratio. We treat $\mu$ and $\beta$ as the design parameters. 
From the functional form, and also intuitively, we know that the free energy, or the associated stresses, should increase with increasing shear modulus so we can pass it to the monotonically increasing branch of ISNN. However, we assume we cannot say anything with certainty about how $\beta$ affects the potential. Therefore, we add it to the free branch of ISNN. Note that in existing pICNN architectures all design parameters can only be passed to the free branch which might result in poor extrapolation performance for the design parameters. To generate the dataset, we generate 500 samples of $\mathbf{F}$ from the deformation gradient space, bounded by $\Delta$ around the undeformed configuration ($\mathbf{F} = \mathbf{I}$), i.e.,
${F}_{ij} \in \delta_{ij} + [-\Delta,\Delta]$
\noindent with $\Delta = 0.2$. We rely on Latin Hypercube Sampling. From Eq. \eqref{psi_blatzko}, we obtain the stresses for different design parameters. For training data, we choose the material parameters from the ranges $\mu \in \{1.0, 7.0\}$ and $\beta \in \{0.125, 2.0\}$. The range chosen for $\beta$ corresponds to a Poisson's ratio $\nu \in \{0.1, 0.4\}$. We therefore generate a dataset $\lbrace (\mathcal{I}_{k}, {\mu}_{i}, {\beta}_{j}) , \mathbf{S}_{k} \rbrace$ with $i,j=1,\ldots,7$ and $k=1,\ldots,500$. Hence, for all the combinations of these inputs, we get $7\times7\times 500$ sets of input data.

\begin{figure}
    \centering
    \includegraphics[scale = 0.35]{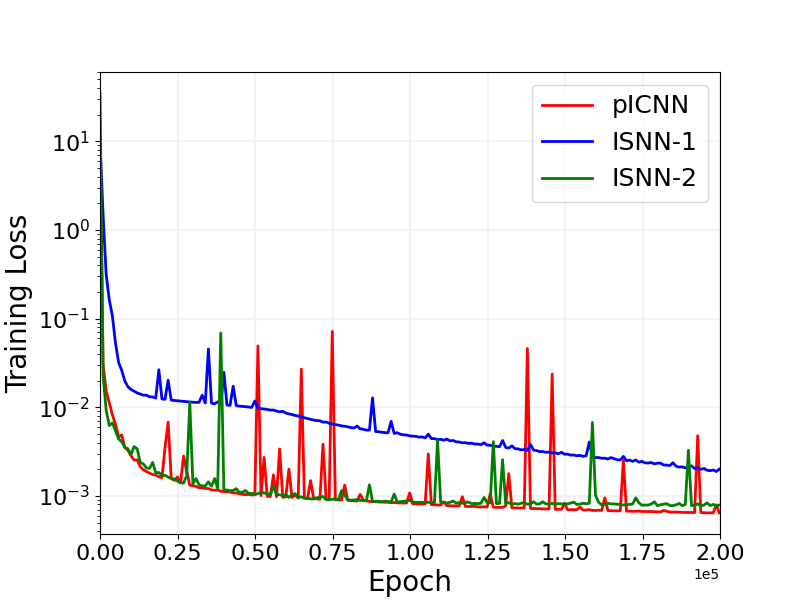}
    \caption{Training loss for different architectures on the isotropic hyperelasticity dataset generated using Eq. \eqref{psi_blatzko}.}
    \label{fig:isoLoss}
\end{figure}

\begin{figure}
    \begin{subfigure}{0.5\linewidth}
    \centering
        \includegraphics[scale=0.35]{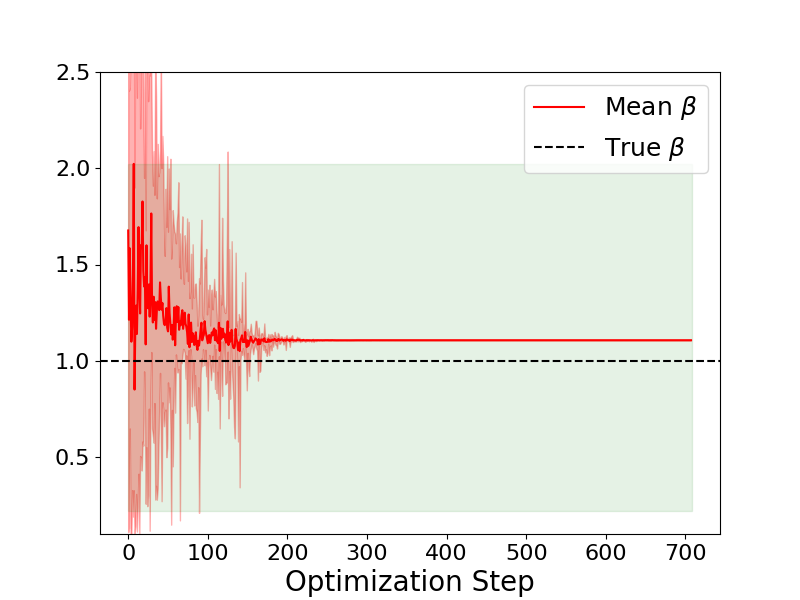}
    \caption{}
    \end{subfigure}
        \begin{subfigure}{0.5\linewidth}
        \centering
        \includegraphics[scale=0.35]{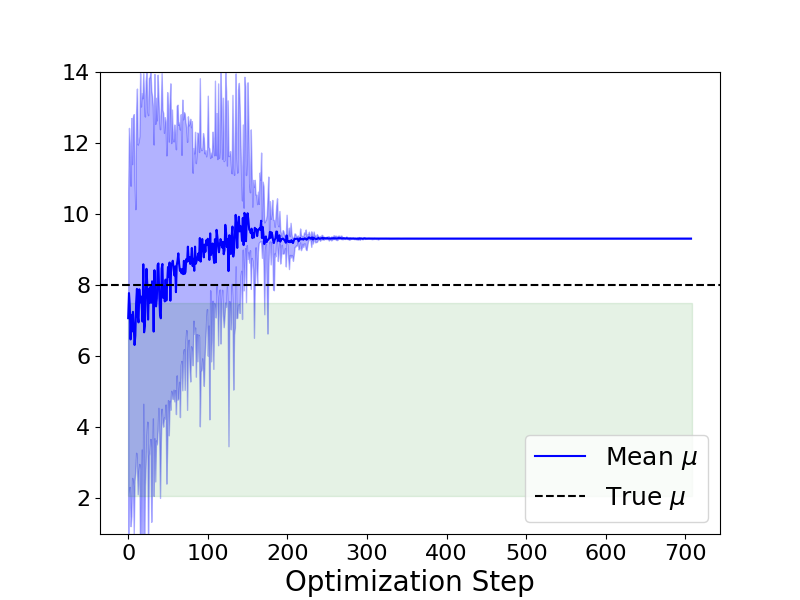}
    \caption{}
    \end{subfigure}
    \caption{The evolution for (a) $\beta$ and (b) $\mu$ for the inverse problem over 10 different initializations using CMA-ES for the parametrized potential trained using pICNN. Here, the solid line indicates the mean value of these parameters with the green-shaded region depicting the training data range.}
    \label{inverse_pICNN_extp}
\end{figure}

\begin{figure}
    \begin{subfigure}{0.5\linewidth}
    \centering
        \includegraphics[scale=0.35]{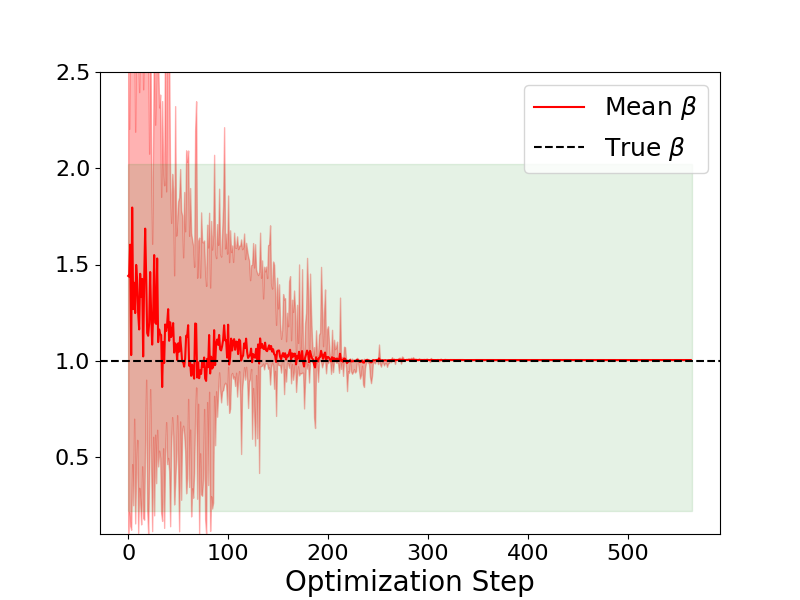}
    \caption{}
    \end{subfigure}
        \begin{subfigure}{0.5\linewidth}
        \centering
        \includegraphics[scale=0.35]{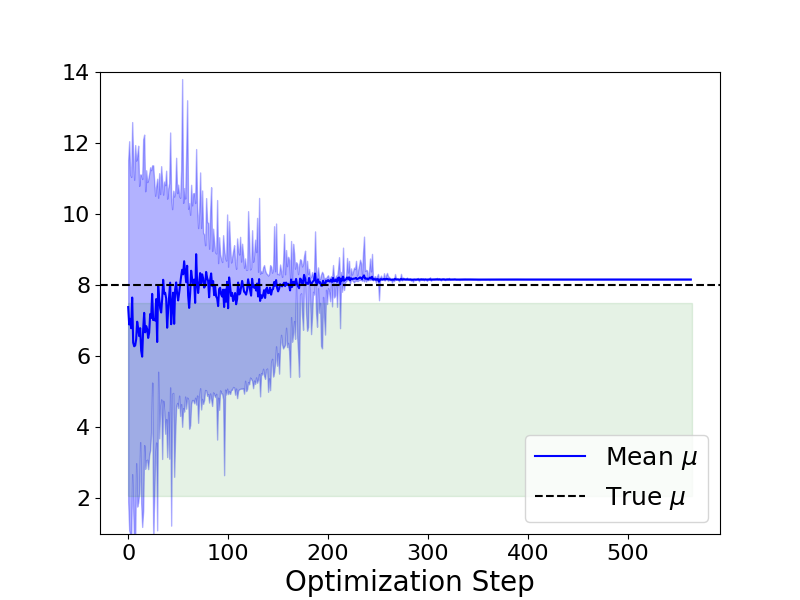}
    \caption{}
    \end{subfigure}
    \caption{The evolution for (a) $\beta$ and (b) $\mu$ for the inverse problem over 10 different initializations using CMA-ES for the parametrized potential trained using ISNN-1. Here, the solid line indicates the mean value of these parameters with the green-shaded region depicting the training data range.}
    \label{inverse_ISNN1_extp}
\end{figure}

\begin{figure}
    \begin{subfigure}{0.5\linewidth}
        \centering
        \includegraphics[scale=0.35]{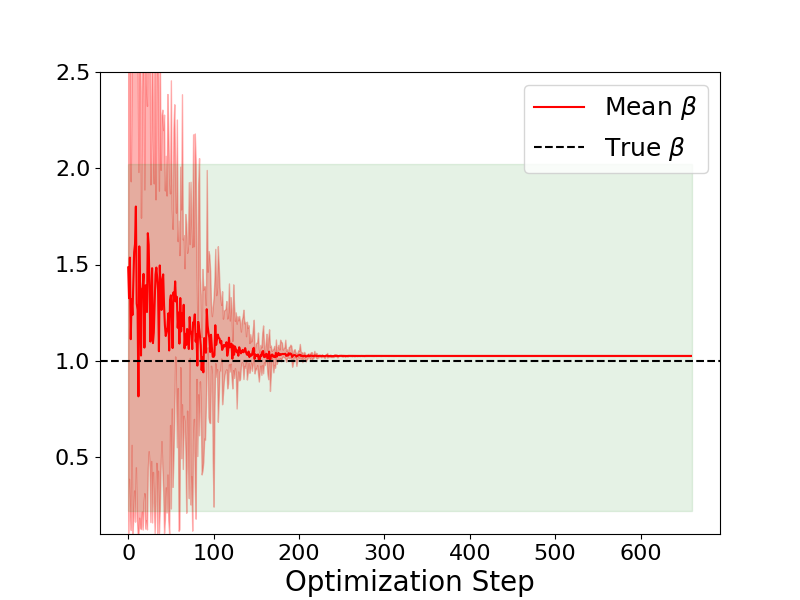}
    \caption{}
    \end{subfigure}
        \begin{subfigure}{0.5\linewidth}
        \centering
        \includegraphics[scale=0.35]{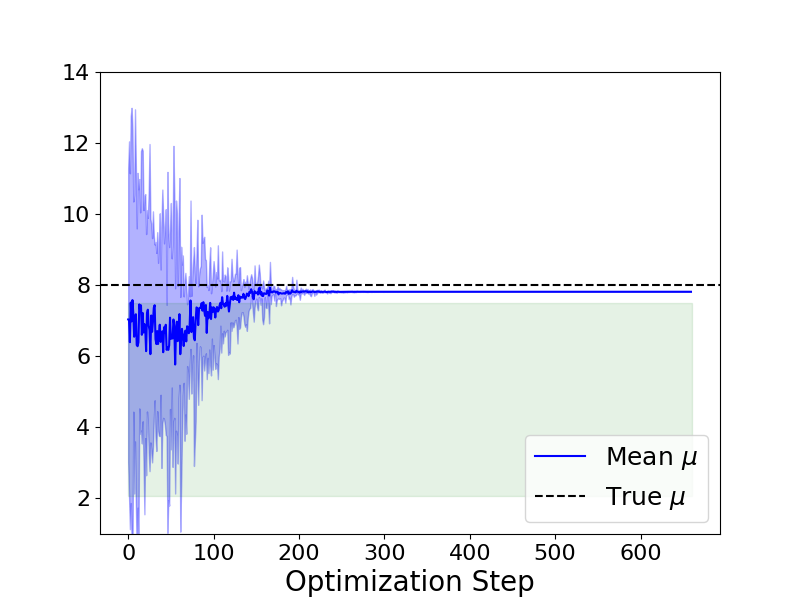}
    \caption{}
    \end{subfigure}
    \caption{The evolution for (a) $\beta$ and (b) $\mu$ for the inverse problem over 10 different initializations using CMA-ES for the parametrized potential trained using ISNN-2. Here, the solid line indicates the mean value of these parameters with the green-shaded region depicting the training data range.}
    \label{inverse_ISNN2_extp}
\end{figure}

\begin{figure}
    \centering
    \includegraphics[scale = 0.35]{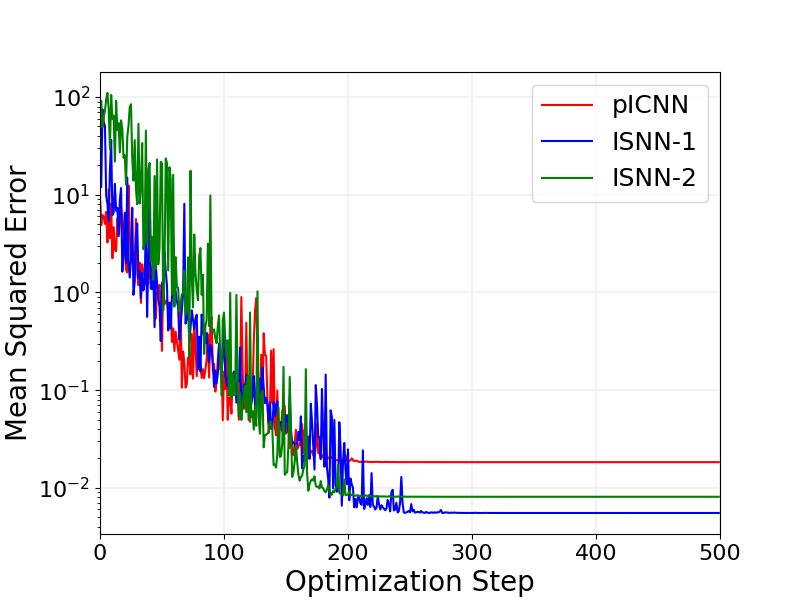}
    \caption{The evolution of the mean squared error (averaged over 10 runs) between the true and predicted stresses over the optimization steps for different NN models.}
    \label{fig:obj_fun_comparison}
\end{figure}

We optimize the neural network parameters over $2\times10^{5}$ epochs. The models were trained on the generated dataset to obtain a parametrized representation of the free energy potential. The training loss for each framework is shown in Figure \ref{fig:isoLoss}. This concludes the forward problem. Now we solve the inverse problem where, using our trained neural networks, we aim to find the design parameters that would result in a desired stress response. To this end, we employ Covariant Matrix Adaptation Evolutionary Strategy (CMA-ES) \cite{cmaes, Hansen16a} to minimize the loss between the target stresses and the ones predicted from the trained NN-based potentials for a given set of design parameters. The results for the inverse problem with both material parameters in the range of the training dataset are presented in Appendix \ref{AppC}. For a more challenging problem, we generate the test target data with extrapolated values for the shear modulus, i.e., we set $\mu=8.0$ and $\beta=1.0$. We ran the inverse problem 10 times for each framework with different initial values for the design parameters and the results for the pICNN model are presented in Figure \ref{inverse_pICNN_extp} whereas Figures \ref{inverse_ISNN1_extp} and \ref{inverse_ISNN2_extp} report the results for ISNN-1 and ISNN-2 respectively. Figure \ref{fig:obj_fun_comparison} shows the evolution of the mean squared error between the true and predicted stresses over the (inverse design) optimization problem. While pICNNs have been employed successfully for such problems \cite{klein2023parametrized, jadoon2024inverse}, we see that ISNNs seem to perform better, especially in extrapolation. To conclude this section, we argue that any sort of prior information that is available about the structural relationship between inputs and outputs should be implicitly incorporated into the neural network.  Doing so should also lead to better performance in the low-data regime. Due to their flexibility ISSNs has shown to extend the set of available tools. 

\subsection{Finite element analysis}
The main goal of modeling constitutive relationships is to use them in structural analysis. For a vast majority of problems in solid mechanics, finite element (FE) analysis is considered the go-to technique for analyzing different structures. For a nonlinear setting in such analyses, the constitutive relationships are called upon at each integration point to get the stress and tangent modulus for the corresponding deformations. Historically, this was done using phenomenological models for free energy and their first and second derivatives for stress and modulus representations, respectively. However, owing to the strides made in the field of constitutive modeling using neural networks, we can now replace these phenomenological models with neural network representations of the free energy \cite{flaschel2023automated, kalina2023fe, Huang2020-xz}. While this serves as a promising alternative for modeling complex material behavior, embedding these networks in commercial finite element software is not always straightforward. This challenge arises from the fact that existing ML libraries rely on automatic differentiation capabilities that allow the users to take derivatives of neural networks. Due to their complexity, it is not straightforward to derive these derivatives manually, especially as their size grows.

\begin{figure}
    \centering
    \includegraphics[scale = 0.7]{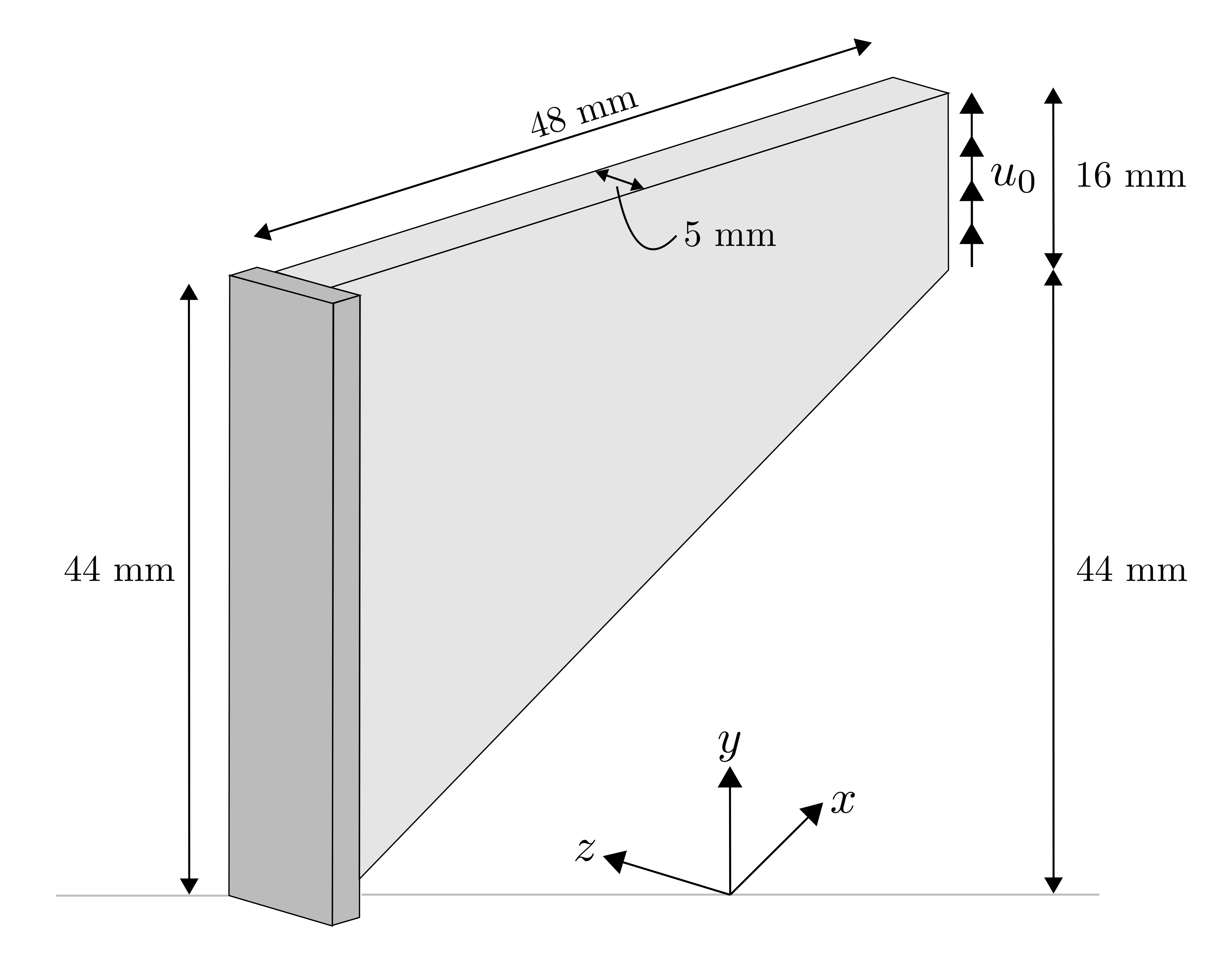}
    \caption{Geometry and boundary conditions for Cook's membrane under consideration for finite element analysis.}
    \label{fig:CooksMembrane}
\end{figure}

Since automatic differentiation is not available in (the most common) commercial solvers practitioners currently rely finite difference schemes to obtain the derivatives of embedded neural networks in these commercial tools \cite{suh2023publicly}. However, the manual derivatives of ISNNs provided in this work, allow for an easy implementation of the developed models into any commercial software. We show this here by implementing hyperelastic constitutive models represented by ISNNs into the FE code \textit{Florence} \cite{Roman2024romeric}. We use manual derivatives to obtain the stresses and the material tangent at the material point.
We analyze a 3D version of the classical Cook's membrane as shown in Figure \ref{fig:CooksMembrane}.

\begin{figure}
    \centering
    \includegraphics[scale = 0.3]{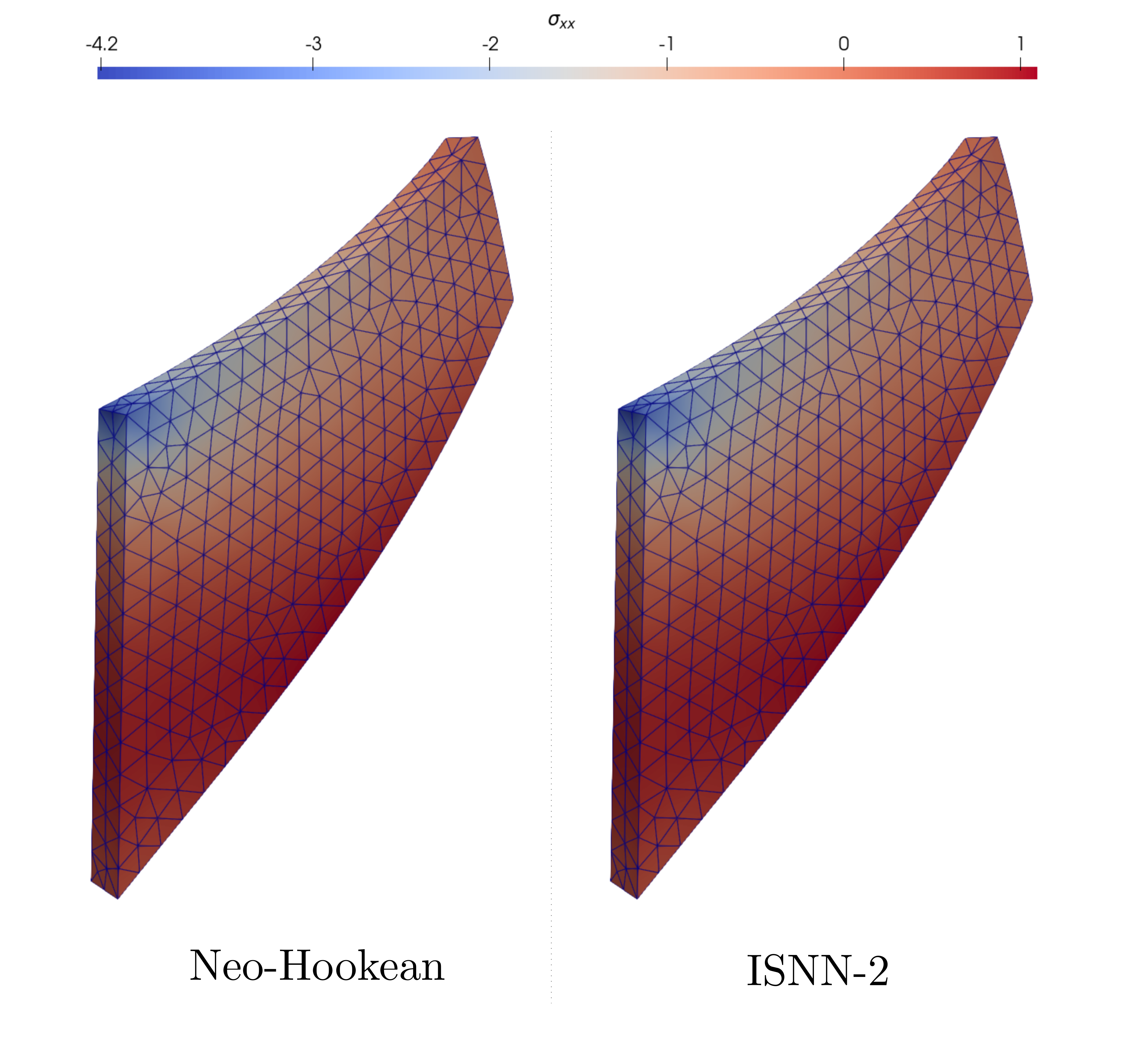}
    \caption{Comparison of the Cauchy stress component $\sigma_{xx}$ for the Neo-hookean free energy from Eq. \eqref{NeohookeanPsi} and trained ISNN.}
    \label{fig:FEstressComparison}
\end{figure}

\begin{figure}
    \centering
    \includegraphics[scale = 0.35]{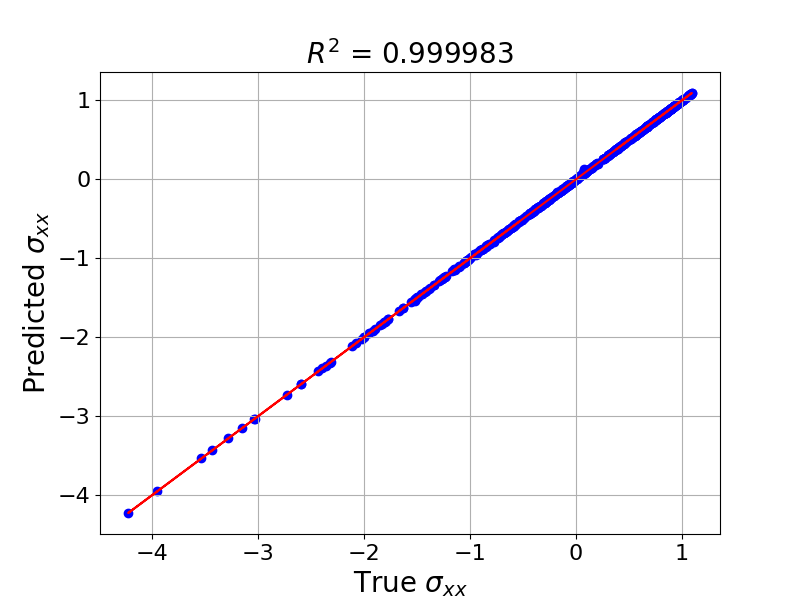}
    \caption{$R^2$ plot of the Cauchy stress component $\sigma_{xx}$ for the Neo-hookean free energy from Eq. \eqref{NeohookeanPsi} denoted as true $\sigma_{xx}$ and trained ISNN denoted as predicted $\sigma_{xx}$.}
    \label{fig:FER2Plot}
\end{figure}

To obtain the constitutive model we first generate a dataset of the form $\lbrace \mathcal{I}, \mathbf{S}\rbrace$ by using the deformation gradients $\mathbf{F}$ sampled in Section \ref{InverseIsotropy} and getting the associated stresses from a Neo-Hookean model described by the free energy potential \cite{ciarlet2021mathematical,bonet1998simple}:
\begin{equation}
\Psi = \frac{1}{2} c_1 (I_1 - 3) - c_1 \log J + \frac{1}{2} c_2 (J -1)^2 \ ,
\label{NeohookeanPsi}
\end{equation}
where $c_1 = 10.0$ and $c_2 = 5.0$. We then train an ISNN-2 using the invariants $\mathcal{I} = \{ \mathcal{I}_1 = \text{tr}\mathbf{C} , \mathcal{I}_2 = \text{tr}\text{Cof}\mathbf{C} , \mathcal{I}_3 = J = \det \mathbf{F} \}$ as inputs with $\mathcal{I}_1$ and $\mathcal{I}_2$ going into the convex, monotonically non-decreasing branch and $J$ going into the convex branch to get a polyconvex representation of the free energy.  In order to validate our results, we run the same FE analysis using the Neo-hookean free energy from Eq. \eqref{NeohookeanPsi}. A comparison of the Cauchy stress component $\sigma_{xx}$ is presented in Figure \ref{fig:FEstressComparison}. Figure \ref{fig:FER2Plot} shows a $R^2$ plot between the $\sigma_{xx}$ over all nodes of the two FE models (ISSN-2 vs. phenomenological model), indicating a good match.

\subsection{Computational cost study}
\begin{figure}
    \centering
    \includegraphics[scale = 0.35]{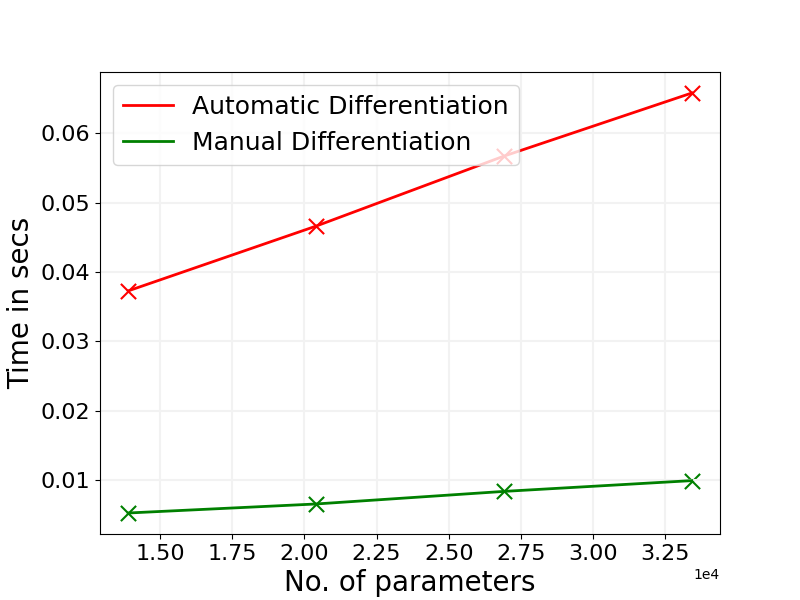}
    \caption{A comparison of time taken (averaged over 10 different initializations of network weights and biases) by AD and MD to evaluate first and second derivatives of ISNN-2.}
    \label{fig:TimeComparison}
\end{figure}

Automatic differentiation (AD) serves as a convenient way of calculating derivatives of the network with respect to the inputs. However, as mentioned earlier, integrating these trained networks into other packages is problematic, and we advocate the use of manual differentiation (MD) using the derivative calculations presented in Appendix \ref{AppA}. To further support our case, we compare the computation time required by AD and MD to calculate the first and second derivatives of the network. With a changing number of total parameters in ISNN-2, we evaluate the derivatives using AD and MD for 10 initializations of the network weights and biases. The average time consumed for each set of parameters is presented in Figure \ref{fig:TimeComparison}. We can clearly see that MD requires less time compared to AD. Some of this time saving can be attributed to the fact that MD can evaluate both the output and the derivatives in a single pass whereas AD requires first the forward pass to get the output and then the backward pass to evaluate the derivatives.

\subsection{Learning structural relationships}
Consider the case where we do not know \textit{a priori} the structural relationship between the inputs and output and we want to find if there exists any. In the example presented here, we focus on discovering polyconvexity from stress-strain data tuples. As explained in Section \ref{InverseIsotropy}, polyconvexity ensures the existence of a minimizer. However, in decoupled multiscale problems where macroscopic response is obtained via homogenization of the microscopic response, it is possible to lose polyconvexity \cite{barchiesi2007loss, braides1994loss, kalina2024neuralnetworksmeetanisotropic} during the homogenization process. Therefore, we now aim to utilize ISNNs to discover if a given dataset comes from a polyconvex potential or not.

\subsubsection{Binary gating mechanism}
One of the main components in such an application of ISNNs is a binary gating mechanism i.e. the gate can only obtain the value 0 or 1. Alternate gating mechanisms have been employed for other problems in solid mechanics \cite{kalina2024neuralnetworksmeetanisotropic}. In our case, we require a binary gate in order to keep the number of inputs and NN parameters to a minimum as will be shown later. One such mechanism can be $L_0$ regularization \cite{louizos2017learning, fuhg2024extreme} which was introduced mainly for the sparsification of neural networks. However, in this work, we present a simple, differentiable binary gating mechanism based on the sigmoid function
\begin{equation}
    \mathcal{S}(g) = \frac{1}{1 + \exp (-\gamma g)} \ ,
    \label{sigmoid}
\end{equation}
where $\gamma > 0$ controls the steepness of the curve. We set $\gamma=1.0$ for our framework. The sigmoid function takes values between 0 and 1. For a binary representation, and while keeping it differentiable, we exploit PyTorch's option to detach operations from the computational graph for the backward pass. Implementation details are presented in Algorithm \ref{algo::BinaryGating}. It can be seen that during the backward pass, the optimizer only sees the sigmoid function regardless of the gate value of 0 or 1 hence allowing us to differentiate through it.     Note that we choose the gate to be 1 if the sigmoid function has a value greater than 0.5 and 0 otherwise. A gate value of 1 indicates that the output can have an arbitrary form with respect to the input and a gate value of 0 imposes that the input-output mapping follows a certain constrained form. Although the optimization network tends to favor the more numerous free or unconstrained parameterizations, we penalize them so they are not preferred unless the data actually represents an unconstrained response. We use $L^p$ regularization to drive $\mathcal{S}(g)$ to zero as will be explained in the next section.

\begin{algorithm2e}

\SetAlgoLined
\SetKwInOut{Input}{Input}
\BlankLine
\Input{$g$}
\BlankLine
$\mathcal{S} = \frac{1}{1 + \exp (-\gamma g)}$ \;
$\Tilde{\mathcal{S}} = \mathcal{S} \text{.detach} $ \;

\eIf{$\mathcal{S}>0.5$}{
$\mathcal{G} = \frac{\mathcal{S}}{\Tilde{\mathcal{S}}}$}
{$\mathcal{G} = \frac{\mathcal{S}}{\Tilde{\mathcal{S}}} - 1.0$}

\Return{$\mathcal{G}$} \;
\caption{Algorithm for binary gating.}
\label{algo::BinaryGating}

\end{algorithm2e}

\subsubsection{Discovering polyconvexity}
\begin{figure}
    \centering
    \includegraphics[scale = 0.5]{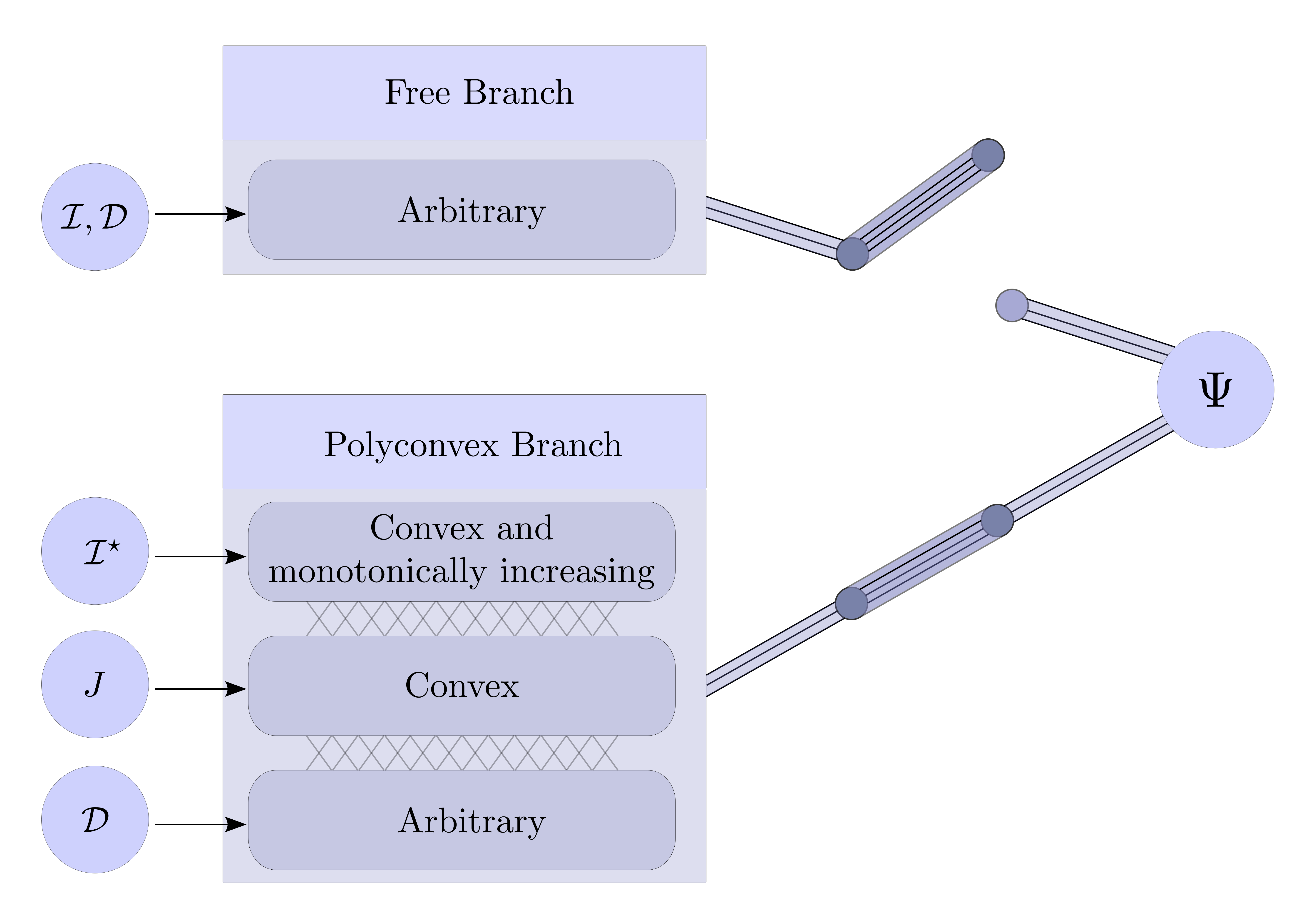}
    \caption{Network architecture for discovering polyconvexity. The schematic shown here assumes the gate value of 0 and therefore, only the polyconvex branch is active.}
    \label{fig:LearningSR}
\end{figure}

We now focus on the neural network architecture used for discovering polyconvexity. Consider a neural network with two independent branches as shown in Figure \ref{fig:LearningSR}. One of these branches forms a polyconvex potential $\Psi_{\text{poly}}$ using ISNN-2 where all the invariants and design parameters go into their respective layers. The other branch takes in all the inputs and outputs a potential $\Psi_{\text{arb}}$ which is unconstrained, i.e., has an arbitrary relationship with the inputs. The potential $\Psi$ then reads:
\begin{equation}
    \Psi = (1-\mathcal{G}) \Psi_{\text{poly}} + \mathcal{G} \Psi_{\text{arb}} \ ,
    \label{polyconvexityPot}
\end{equation}
where $\mathcal{G}$ is the gate value as explained in the previous section. Once again we assume we have access to some dataset $\lbrace (\mathcal{I}, \mathcal{D}), \mathbf{S} \rbrace$. We can train an ISNN to this data with the loss function $\mathcal{L}$:
\begin{equation}
L = \| \mathbf{S} - \hat{\mathbf{S}} \|_2^2 + \varepsilon \| {\mathcal{S}(g)} \|_{p} \ , \quad    \text{with} \quad \| {\mathcal{S}(g)} \|_{p} = \left( \mathcal{S}(g) \right)^\frac{1}{p} \ ,
\end{equation}
where $\varepsilon$ is a penalty parameter that can be tuned for a given problem. In all the examples presented here, we use $\varepsilon=10^{-4}$ and $p=4$. 

Now we can address the need for binary gating. Generally, modifying the loss function in this manner can help drive ${\mathcal{S}(g)}$ to zero in case the actual gate value is 0. However, in case it is 1, doing so might not drive the gate to be binary, i.e., to exactly 1. Consider a gate value between 0 and 1; the network can easily compensate to give the correct output by basically scaling itself. There are cases, c.f. \cite{fuhg2022learning,jadoon2024inverse}, where this turned out to be no issue. However, as we can see from the architecture used here, having a gate value between 0 and 1 is troublesome as that would mean the potential depends on both branches and would result in twice the number of inputs and significantly more weights and biases. Having a binary gating mechanism helps to avoid this since once the model has learned whether the potential is polyconvex or not, we can simply remove all the parameters coming from the other branch. Consequently, we end up with the same number of inputs and reduced network parameters as if we would have known the structural relationship \textit{a priori}. 

\begin{figure}
    \centering
    \includegraphics[scale = 0.25]{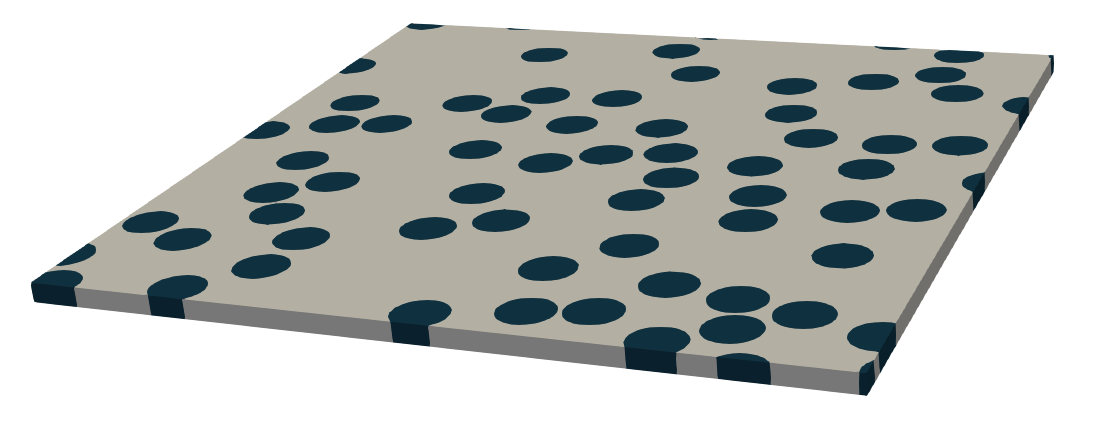}
    \caption{Fiber-reinforced RVE used for generating the stress-strain dataset via computational homogenization.}
    \label{fig:fiberRVE}
\end{figure}

To test our framework, we generate three different datasets for transversely isotropic materials. The first dataset is generated using a polyconvex potential \cite{holzapfel2000new}:
\begin{equation}
    \Psi = c_{1} ( I_{1}- 3 ) + \frac{c_{1}}{c_{2}} (J^{-2 c_{2}}- 1)+ c_{3} \left(\exp( c_{4} (I_{4}-1)^{4}) - 1 \right) \ ,
    \label{dataset_poly}
\end{equation}

where $I_4 \equiv \tr \mathbf{C} \mathbf{N}$ and $\mathbf{N}=\mathbf{n} \otimes \mathbf{n}$, with $\mathbf{n}$ being the preferred direction of anisotropy and $c_i$ are the material or design parameters. Here, we choose $c_1=c_2=c_3=c4=1.0$. The next dataset was generated using a non-polyconvex potential of the form
\begin{equation}
    \Psi = -\frac{c_1}{2}(I_1 - 3) -c_1 \log(J) - \frac{c_2}{2}(J-1)^2 - (I_4 - 1)(c_0 + c_1 \log(J) + c_2(I_4-1)) - \frac{c_0}{2}(\Bar{I}_5 - 1)\ ,
    \label{dataset_nonpoly}
\end{equation}
where $\Bar{I}_5 \equiv \tr \mathbf{C}^2 \mathbf{N}$ and $c_0=c_1=c_2=2.0$. For both of these datasets, we sampled 1000 $\mathbf{F}$ from the deformation gradient space using the methodology mentioned in Section \ref{InverseIsotropy}. The last dataset is obtained via computational homogenization of a fiber-reinforced Representative Volume Element (RVE) shown in Figure \ref{fig:fiberRVE}. The RVE gives a transversely isotropic response in the direction of the fibers and has been shown to lose polyconvexity upon homogenization \cite{kalina2024neuralnetworksmeetanisotropic, jadoon2024inverse}. For transverse isotropy, we have additional invariants such that $\mathcal{I} = \lbrace \mathcal{I}_1, \mathcal{I}_2, \mathcal{I}_3, \mathcal{I}_4 = \tr \mathbf{C} \mathbf{N}, \mathcal{I}_5 = \tr \text{Cof} \mathbf{C} \mathbf{N} \rbrace$ and require to be convex and monotonically increasing with respect to these additional invariants $\mathcal{I}_4$ and $\mathcal{I}_5$. For brevity, we represent all invariants that require convexity and monotonicity with $\mathcal{I}^\star$ i.e. $\mathcal{I}^\star = \lbrace \mathcal{I}_1, \mathcal{I}_2, \mathcal{I}_4, \mathcal{I}_5 \rbrace$. Additionally, this dataset contained parametrized stress-strain tuples with the ratio of the shear moduli of the matrix and fibers, and the volume fraction of fibers chosen as design parameters. 50 sets of these design parameters were sampled along with 840 deformation gradients $\mathbf{F}$ following the sampling technique presented in \cite{KALINA2024116739}. Therefore, the entire dataset contained 840 $\times$ 50 data tuples.

\begin{figure}
    \begin{subfigure}{0.5\linewidth}
        \centering
        \includegraphics[scale=0.35]{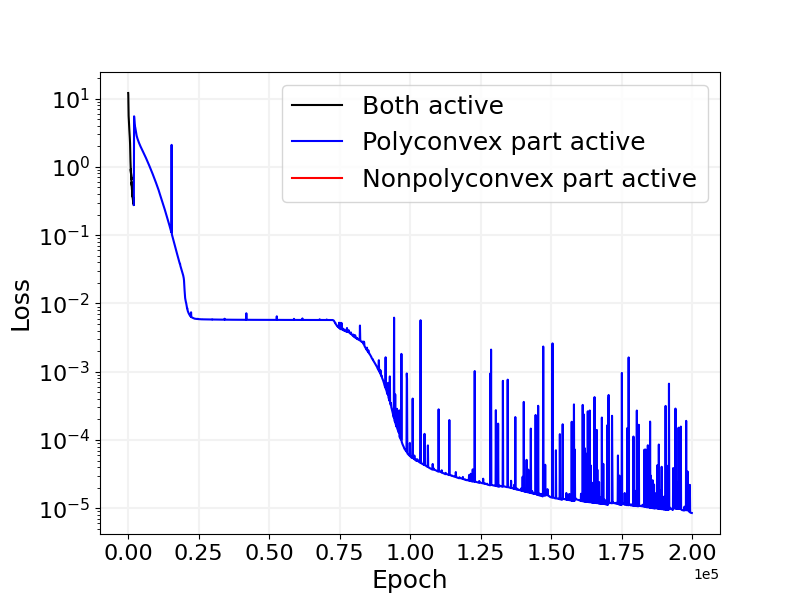}
    \caption{Loss}\label{fig:loss_evol_polyconvex}
    \end{subfigure}
        \begin{subfigure}{0.5\linewidth}
        \centering
        \includegraphics[scale=0.35]{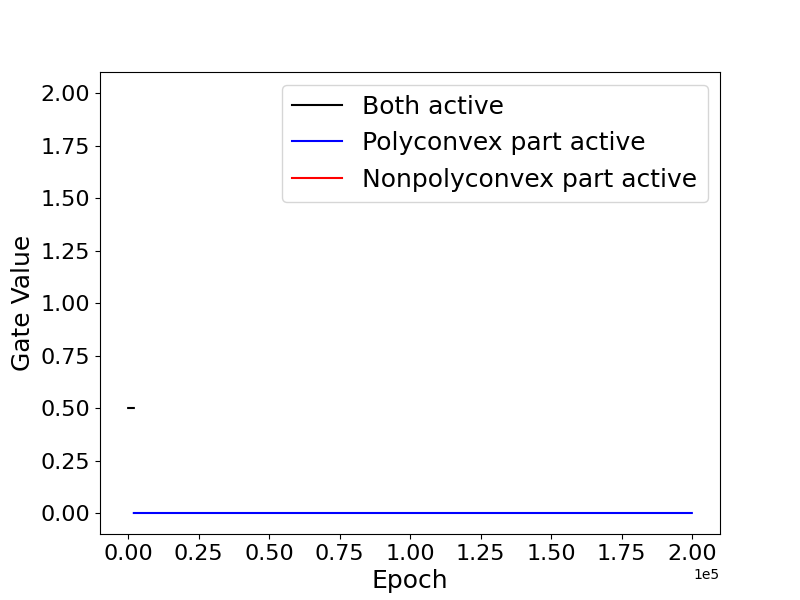}
    \caption{Gate}\label{fig:gate_evol_polyconvex}
    \end{subfigure}
    \caption{(a) Training loss and the (b) evolution of the gate value for the polyconvex dataset from Eq. \eqref{dataset_poly}.}
    \label{results_polyconvex}
\end{figure}

\begin{figure}
    \begin{subfigure}{0.5\linewidth}
        \centering
        \includegraphics[scale=0.35]{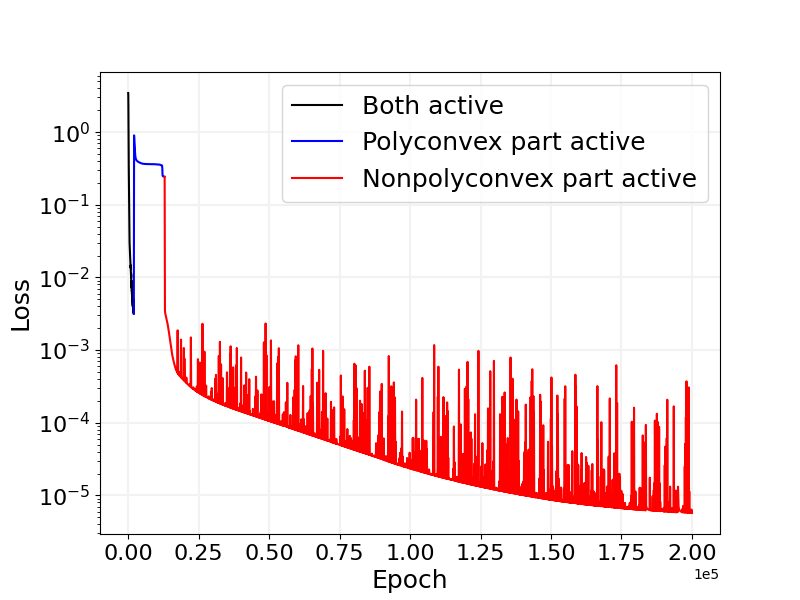}
    \caption{Loss}\label{fig:loss_evol_nonpolyconvex}
    \end{subfigure}
        \begin{subfigure}{0.5\linewidth}
        \centering
        \includegraphics[scale=0.35]{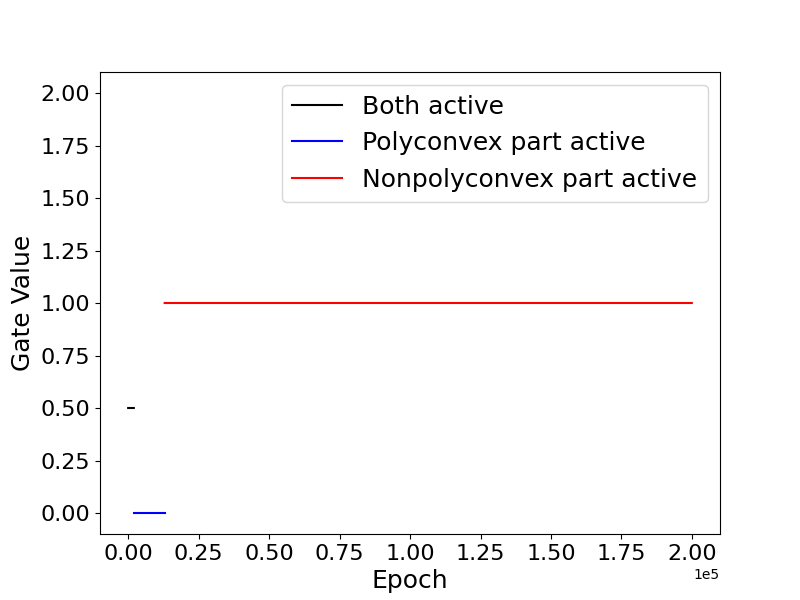}
    \caption{Gate}\label{fig:gate_evol_nonpolyconvex}
    \end{subfigure}
    \caption{(a) Training loss and the (b) evolution of the gate value for the non-polyconvex dataset from Eq. \eqref{dataset_nonpoly}.}
    \label{results_nonpolyconvex}
\end{figure}

\begin{figure}
    \begin{subfigure}{0.5\linewidth}
        \centering
        \includegraphics[scale=0.35]{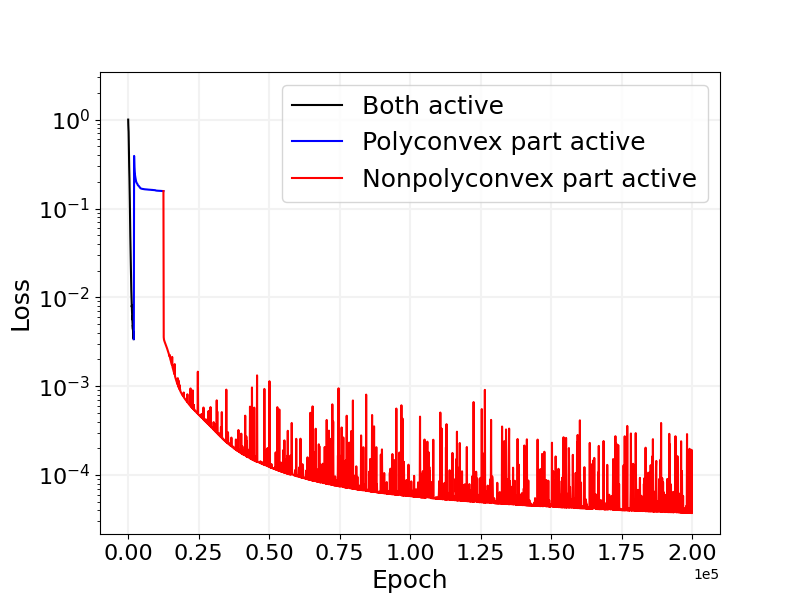}
    \caption{Loss}\label{fig:loss_evol_rve}
    \end{subfigure}
        \begin{subfigure}{0.5\linewidth}
        \centering
        \includegraphics[scale=0.35]{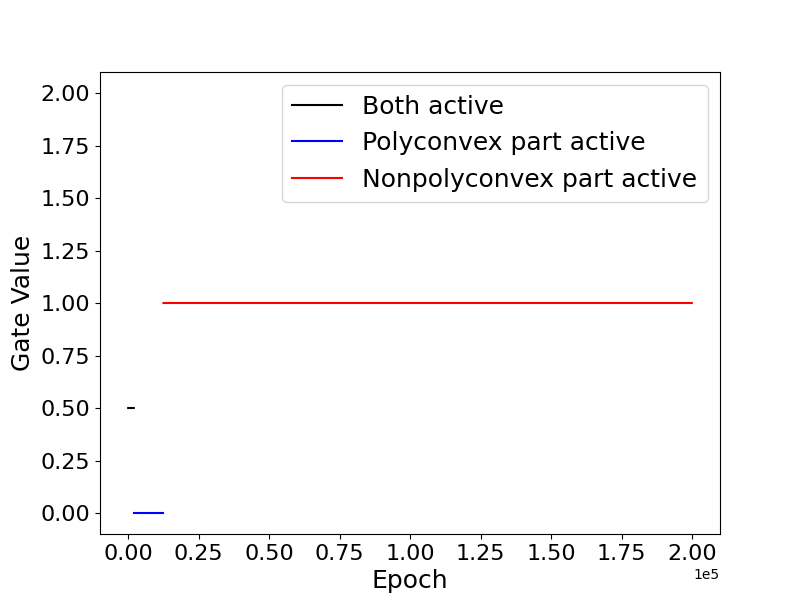}
    \caption{Gate}\label{fig:gate_evol_rve}
    \end{subfigure}
    \caption{(a) Training loss and the (b) evolution of the gate value for the non-polyconvex dataset obtained via computation homogenization of the RVE shown in Figure \ref{fig:fiberRVE}.}
    \label{results_rve}
\end{figure}

Initially, we started the training algorithm of the network by assuming the potential is polyconvex such that $\mathcal{G}=0$. However, in doing so, we observed that if the potential is actually not polyconvex, the neural network struggles to switch the gate. We believe that switching would result in a higher loss since the weights and biases associated with the free branch have not been trained at all. To avoid this issue, we train the first few epochs (1$\%$ of epochs in our case) with both branches active. After the specified number of epochs, we set $g=-0.1$ such that $\mathcal{S}(g)$ is less than 0.5 and the gate value is 0. We run the training process for $2e5$ epochs with a learning rate of $10^{-3}$ for all the weights and biases and $10^{-5}$ for $g$. The training results for the dataset obtained from the polyconvex potential of Eq. \eqref{dataset_poly} are presented in Figure \ref{results_polyconvex} where Figure \ref{fig:loss_evol_polyconvex} shows the training loss and Figure \ref{fig:gate_evol_polyconvex} reports the evolution of the gate value during the training process. Clearly, the network learns that the dataset comes from a polyconvex potential and never activates the free branch. In Figure \ref{results_nonpolyconvex}, the loss and gate evolution are shown for the dataset generated using Eq. \eqref{dataset_nonpoly}. The network learns that it is not possible to fit the data with a polyconvex potential and switches the gate to activate the free branch. Similar results can be seen for the stress-strain tuples coming from the computational homogenization of the fiber-reinforced RVE in Figure \ref{results_rve}, where the network first tries to fit the response using the polyconvex branch but then switches the gate value to 1 thus activating the free branch.

\section{Discussion}
In this study, we presented a novel neural network termed Input Specific Neural Network (ISNN). Two architectures are presented for ISNNs with varying complexity. These networks are tested against other commonly used neural networks to solve various numerical problems. Starting off with two toy problems, we move on to more complex problems in solid mechanics. More specifically, we utilize ISNNs to solve inverse problems in finite strain hyperelasticity, reducing the number of required invariants as compared to the current state-of-the-art and employing them in a finite element setting to act as surrogates for the free energy potential. Manual derivatives are also calculated for each type of ISNN and significant time reduction is observed for manual differentiation as compared to automatic differentiation. We also show that ISNNs can be employed to learn structural relationships, between the inputs and the output. Particularly, we use ISNNs to learn whether a given stress-strain dataset should be modeled with a polyconvex potential or not while using the minimum number of inputs and network parameters. While analytical derivatives of neural networks have been explored before \cite{rodini2022analytical, ratku2022derivatives}, there is not much literature available on the subject. Additionally, the aforementioned references present analytical derivatives of simple feed-forward NNs whereas this work presents an algorithmically efficient way of computing analytical derivatives of a complex network that allows for the passage of information between different branches. The ability to get the derivatives manually for the network allows for a cross-platform implementation of ISNNs where any library might be used for training and inferring from the networks. In future works, we plan to extend ISNNs for constraining multiple outputs and also to employ them for modeling inelastic material behavior \cite{Rosenkranz2023-bd, Rosenkranz2024-dr, jadoon2025automated}. We also plan to incorporate more constraints e.g. concavity constraints with respect to some variables which is directly applicable to problems of thermomechanics \cite{Fuhg2024-fz}. 


\section*{Code and Data}
The codes used in this work for the ISNNs including the training routine and the models' manual derivatives will be made available after the acceptance of this manuscript.

\section*{Acknowledgments}
The authors thank K. A. Kalina and the Kästner-group at TU Dresden for providing the data for the homogenization example.
Sandia National Laboratories is a multimission laboratory managed and operated by National Technology and Engineering Solutions of Sandia, LLC., a wholly owned subsidiary of Honeywell International, Inc., for the U.S.
Department of Energy's National Nuclear Security Administration under contract DE-NA-0003525.
This paper describes objective technical results and analysis.
Any subjective views or opinions that might be expressed in the paper do not necessarily represent the views of the U.S.  Department of Energy or the United States Government.

\clearpage
\appendix
\renewcommand{\theequation}{\thesection.\arabic{equation}} 
\setcounter{equation}{0} 

\section{Derivatives for ISNNs} \label{AppA}

Here, we present explicit derivatives of both ISNN architectures. The derivatives obtained from manual differentiation were compared with those obtained using automatic differentiation in PyTorch. Both calculations resulted in practically identical results barring round-off errors. As an additional numerical experiment, $10^5$ different initializations of network weights and biases were tested and all of them satisfied the imposed constraints.

\subsection*{ISNN-1}

We start by calculating the first derivatives of the output with respect to the input $\mathbf{x}_0$. For this we define
\begin{subequations}
\begin{align}
\mathbb{F}_{0} &= \mathbf{x}_0 \mathbf{W}_0^{[xx]^T} + \mathbf{y}_{H_y} W^{[xy]^T}  + \mathbf{z}_{H_z} W^{[xz]^T} + \mathbf{t}_{H_t} W^{[xt]^T} + \mathbf{b}_0^{[x]} \label{eq:Fx0_1}, \\
\mathbb{F}_{h} &= \mathbf{x}_h \mathbf{W}_h^{[xx]^T} + \mathbf{b}_h^{[x]} , \quad h = 1, \ldots, H_x - 1\,.
\end{align}
\end{subequations}
We know the output of our network would be:
\begin{equation}
\mathbf{x}_{H_x} = \sigma_{mc} \left( \mathbb{F}_{{H_x} - 1} \right) \ .
\label{out_type1}
\end{equation}
Now, we can get the first derivatives as:
\begin{equation}
\begin{aligned}
\frac{\partial \mathbf{x}_{H_x}}{\partial {\mathbf{x}_0}_i} &= \left( \sigma'_{mc} \left(\mathbb{F}_{{H_x}-1} \right)^T\right) \circ \left(\mathbf{W}_{{H_x}-1}^{[xx]} \frac{\partial \mathbf{x}_{{H_x}-1}}{\partial {\mathbf{x}_0}_i} \right) \\
\frac{\partial \mathbf{x}_{H_x - 1}}{\partial {\mathbf{x}_0}_i} &= \left( \sigma'_{mc} \left(\mathbb{F}_{{H_x}-2} \right)^T\right) \circ \left(\mathbf{W}_{{H_x}-2}^{[xx]} \frac{\partial \mathbf{x}_{{H_x}-2}}{\partial {\mathbf{x}_0}_i} \right) \\
    &\vdots\\
\frac{\partial \mathbf{x}_{2}}{\partial {\mathbf{x}_0}_i} &= \left( \sigma'_{mc} \left(\mathbb{F}_{1} \right)^T\right) \circ \left(\mathbf{W}_{1}^{[xx]} \frac{\partial \mathbf{x}_{1}}{\partial {\mathbf{x}_0}_i} \right)\\
    \frac{\partial \mathbf{x}_1}{\partial {\mathbf{x}_0}_i} &= \left( \sigma'_{mc} \left(\mathbb{F}_{0} \right)^T\right) \circ {\mathbf{W}_{0}}_i^{[xx]} \ ,
\end{aligned}
\label{firstDerivativeeqx_type1}
\end{equation}
where $\circ$ represents the Hadamard product and subscript $i$ denotes the input we want to take the derivative to along the corresponding column in the weight matrix. Since $\mathbf{W}_0$ can be negative, the network is not monotonic in the input $\mathbf{x}_0$. Using Eq. \eqref{firstDerivativeeqx_type1}, we can derive the second derivatives as:
\begin{equation}
\begin{aligned}
\frac{\partial^2 \mathbf{x}_{H_x}}{{\partial\mathbf{x}_0}_i \partial{\mathbf{x}_0}_j} &= \left( \sigma''_{mc} \left(\mathbb{F}_{H_x -1} \right)^T\right) \circ \left(\mathbf{W}_{H_x -1}^{[xx]} \frac{\partial \mathbf{x}_{H_x -1}}{\partial{\mathbf{x}_0}_i}\right) \circ \left(\mathbf{W}_{H_x -1}^{[xx]} \frac{\partial \mathbf{x}_{H_x -1}}{\partial{\mathbf{x}_0}_j} \right) \\
&+ \left( \sigma'_{mc} \left(\mathbb{F}_{H_x -1} \right)^T\right) \circ \left(\mathbf{W}_{H_x -1}^{[xx]} \frac{\partial^2 \mathbf{x}_{H_x - 1}}{\partial{\mathbf{x}_0}_i \partial{\mathbf{x}_0}_j} \right) \\
\frac{\partial^2 \mathbf{x}_{H_x - 1}}{{\partial\mathbf{x}_0}_i \partial{\mathbf{x}_0}_j} &= \left( \sigma''_{mc} \left(\mathbb{F}_{H_x -2} \right)^T\right) \circ \left(\mathbf{W}_{H_x -2}^{[xx]} \frac{\partial \mathbf{x}_{H_x -2}}{\partial{\mathbf{x}_0}_i}\right) \circ \left(\mathbf{W}_{H_x -1}^{[xx]} \frac{\partial \mathbf{x}_{H_x -2}}{\partial{\mathbf{x}_0}_j} \right) \\
&+ \left( \sigma'_{mc} \left(\mathbb{F}_{H_x -2} \right)^T\right) \circ \left(\mathbf{W}_{H_x -2}^{[xx]} \frac{\partial^2 \mathbf{x}_{H_x-2}}{\partial{\mathbf{x}_0}_i \partial{\mathbf{x}_0}_j} \right) \\
    &\vdots\\
\frac{\partial^2 \mathbf{x}_{2}}{{\partial\mathbf{x}_0}_i \partial{\mathbf{x}_0}_j} &= \left( \sigma''_{mc} \left(\mathbb{F}_{1} \right)^T\right) \circ \left(\mathbf{W}_{1}^{[xx]} \frac{\partial \mathbf{x}_{1}}{\partial{\mathbf{x}_0}_i}\right) \circ \left(\mathbf{W}_{1}^{[xx]} \frac{\partial \mathbf{x}_{1}}{\partial{\mathbf{x}_0}_j} \right) \\
&+ \left( \sigma'_{mc} \left(\mathbb{F}_{1} \right)^T\right) \circ \left(\mathbf{W}_{1}^{[xx]} \frac{\partial^2 \mathbf{x}_{1}}{\partial{\mathbf{x}_0}_i \partial{\mathbf{x}_0}_j} \right) \\
    \frac{\partial^2 \mathbf{x}_1}{{\partial\mathbf{x}_0}_i \partial{\mathbf{x}_0}_j} &= \left( \sigma''_{mc} \left(\mathbb{F}_{0} \right)^T\right) \circ {\mathbf{W}_{0}}_i^{[xx]} \circ {\mathbf{W}_{0}}_j^{[xx]} \ .
\end{aligned}
\label{secondDerivativeeqx_type1}
\end{equation}
While it is convenient to write the derivatives in terms of components of the Hessian, we use a more general representation to show the proof of convexity. Consider the Hessian of the first layer, which we can write as:
\begin{equation}
        \frac{\partial^2 \mathbf{x}_1}{{\partial\mathbf{x}_0}^2} = {{\mathbf{W}_{0}}^{[xx]}}^T \cdot \left( \sigma''_{mc} \left(\mathbb{F}_{0} \right)^T \cdot \mathbf{J}\right) \cdot{\mathbf{W}_{0}}^{[xx]} \ , \label{temp_1}
\end{equation}
where $\cdot$ represents tensor multiplication and $\mathbf{J}$ is a row vector of ones with the same size as $\sigma''_{mc} \left(\mathbb{F}_{0}\right)$. To prove positive semidefiniteness, and in turn convexity, we have to show:
\begin{equation}
        \mathbf{v}^T \cdot \frac{\partial^2 \mathbf{x}_1}{{\partial\mathbf{x}_0}^2} \cdot  \mathbf{v} \geq 0 \ ,
\end{equation}
for any $\mathbf{v}$. Using Eq. \eqref{temp_1}, this becomes:
\begin{equation}
\begin{aligned}
        \mathbf{v}^T \cdot {{\mathbf{W}_{0}}^{[xx]}}^T\cdot\left( \sigma''_{mc} \left(\mathbb{F}_{0} \right)^T\cdot \mathbf{J}\right) \cdot{\mathbf{W}_{0}}^{[xx]} \cdot  \mathbf{v} &\geq 0 \ \\
        ({{\mathbf{W}_{0}}^{[xx]}} \cdot \mathbf{v})^T\cdot\left( \sigma''_{mc} \left(\mathbb{F}_{0} \right)^T\cdot \mathbf{J}\right) \cdot({\mathbf{W}_{0}}^{[xx]} \cdot  \mathbf{v}) &\geq 0 \ .
\end{aligned}
\end{equation}
Now, setting $\mathbf{u} = \mathbf{v}\cdot\mathbf{W}_0$, we end up with the condition:
\begin{equation}
        \mathbf{u}^T \cdot\left( \sigma''_{mc} \left(\mathbb{F}_{0} \right)^T\cdot \mathbf{J}\right) \cdot  \mathbf{u} \geq 0 \ 
\end{equation}
Therefore, we only need to prove the positive semidefiniteness of $(\sigma''_{mc} \left(\mathbb{F}_{0} \right)^T\cdot \mathbf{J})$ which gives us a matrix with identical columns. All the eigenvalues of this rank-one matrix are zero, except for one. This non-zero eigenvalue is the sum of all elements in a column. Since $\sigma_{mc}$ is constrained to be convex, all the entries in the columns are non-negative resulting in a non-negative eigenvalue and a positive semidefinite Hessian. In subsequent layers, e.g., 
\begin{equation}
\frac{\partial^2 \mathbf{x}_{2}}{{\partial\mathbf{x}_0}_i \partial{\mathbf{x}_0}_j} = \left( \sigma''_{mc} \left(\mathbb{F}_{1} \right)^T\right) \circ \left(\mathbf{W}_{1}^{[xx]} \frac{\partial \mathbf{x}_{1}}{\partial{\mathbf{x}_0}_i}\right) \circ \left(\mathbf{W}_{1}^{[xx]} \frac{\partial \mathbf{x}_{1}}{\partial{\mathbf{x}_0}_j} \right) 
+ \left( \sigma'_{mc} \left(\mathbb{F}_{1} \right)^T\right) \circ \left(\mathbf{W}_{1}^{[xx]} \frac{\partial^2 \mathbf{x}_{1}}{\partial{\mathbf{x}_0}_i \partial{\mathbf{x}_0}_j} \right)
\end{equation}
the positive definiteness of the first term can be established using similar reasoning. In the second term, the positive semidefinite Hessian of the last layer is multiplied by only positive values and thereby retains its positive definiteness. Since the sum of two positive semidefinite matrices is also positive semidefinite we can prove that all subsequent Hessians are also positive semidefinite.



Now we calculate the derivatives with respect to the inputs $\mathbf{y}_0$. Letting:
\begin{equation}
\mathbb{G}_{h} = \left(\mathbf{y}_h  \mathbf{W}_h^{[yy]^T} + \mathbf{b}_h^{[y]} \right) , \quad h = 0, \ldots, H_y - 1 \ ,
\end{equation}
we get the first derivatives as:
\begin{equation}
\begin{aligned}
\frac{\partial \mathbf{x}_{H_x}}{\partial {\mathbf{y}_0}_i} &= \left( \sigma'_{mc} \left(\mathbb{F}_{{H_x}-1} \right)^T\right) \circ \left(\mathbf{W}_{{H_x}-1}^{[xx]} \frac{\partial \mathbf{x}_{{H_x}-1}}{\partial {\mathbf{y}_0}_i} \right) \\
    &\vdots\\
\frac{\partial \mathbf{x}_2}{\partial {\mathbf{y}_0}_i} &= \left( \sigma'_{mc} \left(\mathbb{F}_{1} \right)^T\right) \circ \left({\mathbf{W}_{1}}^{[xx]} \frac{\partial \mathbf{x}_{1}}{\partial {\mathbf{y}_0}_i}\right)\\
\frac{\partial \mathbf{x}_1}{\partial {\mathbf{y}_0}_i} &= \left( \sigma'_{mc} \left(\mathbb{F}_{0} \right)^T\right) \circ \left({W}^{[xy]} \frac{\partial \mathbf{y}_{{H_y}}}{\partial {\mathbf{y}_0}_i}\right) \ ,
\end{aligned}
\label{firstDerivativeeqxy_type1}
\end{equation}
and,
\begin{equation}
\begin{aligned}
\frac{\partial \mathbf{y}_{H_y}}{\partial {\mathbf{y}_0}_i} &= \left( \sigma'_{mc} \left(\mathbb{G}_{{H_y}-1} \right)^T\right) \circ \left(\mathbf{W}_{{H_y}-1}^{[yy]} \frac{\partial \mathbf{y}_{{H_y}-1}}{\partial {\mathbf{y}_0}_i} \right) \\
\frac{\partial \mathbf{y}_{H_y - 1}}{\partial {\mathbf{y}_0}_i} &= \left( \sigma'_{mc} \left(\mathbb{G}_{{H_y}-2} \right)^T\right) \circ \left(\mathbf{W}_{{H_y}-2}^{[yy]} \frac{\partial \mathbf{y}_{{H_y}-2}}{\partial {\mathbf{y}_0}_i} \right) \\
    &\vdots\\
\frac{\partial \mathbf{y}_{2}}{\partial {\mathbf{y}_0}_i} &= \left( \sigma'_{mc} \left(\mathbb{G}_{1} \right)^T\right) \circ \left(\mathbf{W}_{1}^{[yy]} \frac{\partial \mathbf{y}_{1}}{\partial {\mathbf{y}_0}_i} \right)\\
    \frac{\partial \mathbf{y}_1}{\partial {\mathbf{y}_0}_i} &= \left( \sigma'_{mc} \left(\mathbb{G}_{0} \right)^T\right) \circ {\mathbf{W}_{0}}_i^{[yy]} \ .
\end{aligned}
\label{firstDerivativeeqy_type1}
\end{equation}
We can see that for the network to be monotonically increasing with respect to $\mathbf{y}_0$, the weights $W^{[yy]}$ and $W^{[xy]}$ need to be positive otherwise the first derivative could have a negative value. The weights $W^{[xx]}_h$ are already positive for $h \geq 1$. Also, $\sigma_{mc}$ is chosen to be a monotonically increasing function so that $\sigma'_{mc}$ is always positive. For the second derivatives we have:
\begin{equation}
\begin{aligned}
\frac{\partial^2 \mathbf{x}_{H_x}}{\partial{\mathbf{y}_0}_i \partial{\mathbf{y}_0}_j} &= \left( \sigma''_{mc} \left(\mathbb{F}_{H_x -1} \right)^T\right) \circ \left(\mathbf{W}_{H_x -1}^{[xx]} \frac{\partial \mathbf{x}_{H_x -1}}{\partial{\mathbf{y}_0}_i}\right) \circ \left(\mathbf{W}_{H_x -1}^{[xx]} \frac{\partial \mathbf{x}_{H_x -1}}{\partial{\mathbf{y}_0}_j} \right) \\
&+ \left( \sigma'_{mc} \left(\mathbb{F}_{H_x -1} \right)^T\right) \circ \left(\mathbf{W}_{H_x -1}^{[xx]} \frac{\partial^2 \mathbf{x}_{H_x - 1}}{\partial{\mathbf{y}_0}_i \partial{\mathbf{y}_0}_j} \right) \\
    &\vdots\\
\frac{\partial^2 \mathbf{x}_{2}}{\partial{\mathbf{y}_0}_i \partial{\mathbf{y}_0}_j} &= \left( \sigma''_{mc} \left(\mathbb{F}_{1} \right)^T\right) \circ \left(\mathbf{W}_{1}^{[xx]} \frac{\partial \mathbf{x}_{1}}{\partial{\mathbf{y}_0}_i}\right) \circ \left(\mathbf{W}_{1}^{[xx]} \frac{\partial \mathbf{x}_{1}}{\partial{\mathbf{y}_0}_j} \right) \\
&+ \left( \sigma'_{mc} \left(\mathbb{F}_{1} \right)^T\right) \circ \left(\mathbf{W}_{1}^{[xx]} \frac{\partial^2 \mathbf{x}_{1}}{\partial{\mathbf{y}_0}_i \partial{\mathbf{y}_0}_j} \right) \\
    \frac{\partial^2 \mathbf{x}_1}{\partial{\mathbf{y}_0}_i \partial{\mathbf{y}_0}_j} &= \left( \sigma''_{mc} \left(\mathbb{F}_{0} \right)^T\right) \circ \left(W^{[xy]} \frac{\partial \mathbf{y}_{H_y}}{\partial{\mathbf{y}_0}_i}\right) \circ \left(W^{[xy]} \frac{\partial \mathbf{y}_{H_y}}{\partial{\mathbf{y}_0}_j} \right) \\
&+ \left( \sigma'_{mc} \left(\mathbb{F}_{0} \right)^T\right) \circ \left(W^{[xy]} \frac{\partial^2 \mathbf{y}_{H_y}}{\partial{\mathbf{y}_0}_i \partial{\mathbf{y}_0}_j} \right) \ , \\
\end{aligned}
\label{secondDerivativeeqxy_type1}
\end{equation}
and,
\begin{equation}
\begin{aligned}
\frac{\partial^2 \mathbf{y}_{H_y}}{\partial{\mathbf{y}_0}_i \partial{\mathbf{y}_0}_j} &= \left( \sigma''_{mc} \left(\mathbb{G}_{H_y -1} \right)^T\right) \circ \left(\mathbf{W}_{H_y -1}^{[yy]} \frac{\partial \mathbf{y}_{H_y -1}}{\partial{\mathbf{y}_0}_i}\right) \circ \left(\mathbf{W}_{H_y -1}^{[yy]} \frac{\partial \mathbf{y}_{H_y -1}}{\partial{\mathbf{y}_0}_j} \right) \\
&+ \left( \sigma'_{mc} \left(\mathbb{G}_{H_y -1} \right)^T\right) \circ \left(\mathbf{W}_{H_y -1}^{[yy]} \frac{\partial^2 \mathbf{y}_{H_y - 1}}{\partial{\mathbf{y}_0}_i \partial{\mathbf{y}_0}_j} \right) \\
    &\vdots\\
\frac{\partial^2 \mathbf{y}_{2}}{\partial{\mathbf{y}_0}_i \partial{\mathbf{y}_0}_j} &= \left( \sigma''_{mc} \left(\mathbb{G}_{1} \right)^T\right) \circ \left(\mathbf{W}_{1}^{[yy]} \frac{\partial \mathbf{y}_{1}}{\partial{\mathbf{y}_0}_i}\right) \circ \left(\mathbf{W}_{1}^{[yy]} \frac{\partial \mathbf{y}_{1}}{\partial{\mathbf{y}_0}_j} \right) \\
&+ \left( \sigma'_{mc} \left(\mathbb{G}_{1} \right)^T\right) \circ \left(\mathbf{W}_{1}^{[yy]} \frac{\partial^2 \mathbf{y}_{1}}{\partial{\mathbf{y}_0}_i \partial{\mathbf{y}_0}_j} \right) \\
    \frac{\partial^2 \mathbf{y}_1}{\partial{\mathbf{y}_0}_i \partial{\mathbf{y}_0}_j} &= \left( \sigma''_{mc} \left(\mathbb{G}_{0} \right)^T\right) \circ {\mathbf{W}_{0}}_i^{[yy]} \circ {\mathbf{W}_{0}}_j^{[yy]} \ .
\end{aligned}
\label{secondDerivativeeqy_type1}
\end{equation}


The proof for convexity of the network with respect to $\mathbf{y}_0$ follows the same ideas presented for convexity with respect to $\mathbf{x}_0$.

Similarly, we can prove that for the inputs $\mathbf{t}_0$, we only need to have a positive monotonically non-decreasing activation function because the first derivatives should be positive. Lastly, since we can be arbitrary with respect to the inputs $\mathbf{z}_0$, there are no constraints on the weights or the activation functions associated with input $\mathbf{z}_0$.

\subsection*{ISNN-2}

In order to calculate the first derivatives of the output with respect to the input $\mathbf{x}_0$, we let:
\begin{subequations}
\begin{align}
\mathbb{F}_{0} &= \mathbf{x}_0 \mathbf{W}_0^{[xx]^T} + \mathbf{y}_0 \mathbf{W}_0^{[xy]^T}  + \mathbf{z}_0 \mathbf{W}_0^{[xz]^T} + \mathbf{t}_0 \mathbf{W}_0^{[xt]^T} + \mathbf{b}_0^{[x]} \label{eq:Fx0_2} \ , \\
\mathbb{F}_{h} &= \mathbf{x}_h \mathbf{W}_h^{[xx]^T} +  \mathbf{x}_0 \mathbf{W}_h^{[xx_{0}]^T} + \mathbf{y}_h \mathbf{W}_h^{[xy]^T}  + \mathbf{z}_h \mathbf{W}_h^{[xz]^T} + \mathbf{t}_h \mathbf{W}_h^{[xt]^T} + \mathbf{b}_h^{[x]} , \quad h = 1, \ldots, H - 1 \ .
\end{align}
\end{subequations}
From this, the output of our network reads
\begin{equation}
\mathbf{x}_{H} = \sigma_{mc} \left( \mathbb{F}_{H-1} \right) \ ,
\label{x_eq_inF_2}
\end{equation}
with the derivatives
\begin{equation}
\begin{aligned}
\frac{\partial \mathbf{x}_{H}}{\partial {\mathbf{x}_0}_i} &= \left( \sigma'_{mc} \left(\mathbb{F}_{H-1} \right)^T \right) \circ \left(\mathbf{W}_{H-1}^{[xx]} \frac{\partial \mathbf{x}_{H-1}}{\partial {\mathbf{x}_0}_i} + {\mathbf{W}_{H-1}^{[xx_{0}]}}_i \right) \\
\frac{\partial \mathbf{x}_{H-1}}{\partial {\mathbf{x}_0}_i} &= \left( \sigma'_{mc} \left(\mathbb{F}_{H-2} \right)^T \right) \circ \left(\mathbf{W}_{H-2}^{[xx]} \frac{\partial \mathbf{x}_{H-2}}{\partial {\mathbf{x}_0}_i} + {\mathbf{W}_{H-2}^{[xx_{0}]}}_i \right)\\
    &\vdots\\
\frac{\partial \mathbf{x}_{2}}{\partial {\mathbf{x}_0}_i} &= \left( \sigma'_{mc} \left(\mathbb{F}_{1} \right)^T \right) \circ \left(\mathbf{W}_{1}^{[xx]} \frac{\partial \mathbf{x}_{1}}{\partial {\mathbf{x}_0}_i} + \mathbf{W}_{1i}^{[xx_{0}]} \right)\\
    \frac{\partial \mathbf{x}_1}{\partial {\mathbf{x}_0}_i} &= \left( \sigma'_{mc} \left(\mathbb{F}_{0} \right)^T \right) \circ {\mathbf{W}_{0i}^{[xx]}} \ .
\end{aligned}
\label{firstDerivativeeqx_type2}
\end{equation}
Similarly, we get the second derivatives with
\begin{equation}
\begin{aligned}
\frac{\partial^2 \mathbf{x}_{H}}{{\partial\mathbf{x}_0}_i \partial{\mathbf{x}_0}_j} &= \left( \sigma''_{mc} \left(\mathbb{F}_{H-1} \right)^T \right) \circ \left(\mathbf{W}_{H-1}^{[xx]} \frac{\partial \mathbf{x}_{H-1}}{\partial{\mathbf{x}_0}_i} + {\mathbf{W}_{H-1}^{[xx_{0}]}}_i \right) \circ \left(\mathbf{W}_{H-1}^{[xx]} \frac{\partial \mathbf{x}_{H-1}}{\partial{\mathbf{x}_0}_j} + {\mathbf{W}_{H-1}^{[xx_{0}]}}_j \right) \\
&+ \left( \sigma'_{mc} \left(\mathbb{F}_{H-1} \right)^T \right) \circ \left(\mathbf{W}_{H-1}^{[xx]} \frac{\partial^2 \mathbf{x}_{H-1}}{\partial{\mathbf{x}_0}_i \partial{\mathbf{x}_0}_j} \right) \\
\frac{\partial^2 \mathbf{x}_{H-1}}{{\partial\mathbf{x}_0}_i \partial{\mathbf{x}_0}_j} &= \left( \sigma''_{mc} \left(\mathbb{F}_{H-2} \right)^T \right) \circ \left(\mathbf{W}_{H-2}^{[xx]} \frac{\partial \mathbf{x}_{H-2}}{\partial{\mathbf{x}_0}_i} + {\mathbf{W}_{H-2}^{[xx_{0}]}}_i \right) \circ \left(\mathbf{W}_{H-2}^{[xx]} \frac{\partial \mathbf{x}_{H-2}}{\partial{\mathbf{x}_0}_j} + {\mathbf{W}_{H-2}^{[xx_{0}]}}_j \right)\\
&+ \left( \sigma'_{mc} \left(\mathbb{F}_{H-2} \right)^T \right) \circ \left(\mathbf{W}_{H-2}^{[xx]} \frac{\partial^2 \mathbf{x}_{H-2}}{\partial{\mathbf{x}_0}_i \partial{\mathbf{x}_0}_j} \right) \\
    &\vdots\\
\frac{\partial^2 \mathbf{x}_{2}}{{\partial\mathbf{x}_0}_i \partial{\mathbf{x}_0}_j} &= \left( \sigma''_{mc} \left(\mathbb{F}_{1} \right)^T \right) \circ \left(\mathbf{W}_{1}^{[xx]} \frac{\partial \mathbf{x}_{1}}{\partial{\mathbf{x}_0}_i} + \mathbf{W}_{1i}^{[xx_{0}]} \right) \circ \left(\mathbf{W}_{1}^{[xx]} \frac{\partial \mathbf{x}_{1}}{\partial{\mathbf{x}_0}_j} + \mathbf{W}_{1j}^{[xx_{0}]} \right) \\
&+ \left( \sigma'_{mc} \left(\mathbb{F}_{1} \right)^T \right) \circ \left(\mathbf{W}_{1}^{[xx]} \frac{\partial^2 \mathbf{x}_{1}}{{\mathbf{x}_0}_i {\mathbf{x}_0}_j}\right)\\
    \frac{\partial^2 \mathbf{x}_1}{{\partial\mathbf{x}_0}_i \partial{\mathbf{x}_0}_j} &= \left( \sigma''_{mc} \left(\mathbb{F}_{0} \right)^T \right) \circ \mathbf{W}_{0i}^{[xx]} \circ \mathbf{W}_{0j}^{[xx]} \ .
\end{aligned}
\label{secondDerivativeeqx_2}
\end{equation}
We now calculate the derivatives of the neural network output with respect to the inputs $\mathbf{y}_0$ by first letting:
\begin{equation}
\mathbb{G}_h = \mathbf{y}_h \mathbf{W}_h^{yy} + b_h^{yy} , \quad h = 0, \ldots, H - 2 \ .
\end{equation}
Then, similar to the proof for ISNN-1, we can evaluate derivatives of the output with respect to $\mathbf{y}_0$ as:
\begin{equation}
\frac{\partial \mathbf{x}_{H}}{\partial {\mathbf{y}_0}_i} = \left( \sigma'_{mc} \left(\mathbb{F}_{H-1} \right)^T \right) \circ \left(\mathbf{W}_{H-1}^{[xx]} \frac{\partial \mathbf{x}_{H-1}}{\partial {\mathbf{y}_0}_i} + \mathbf{W}_{H-1}^{[xy]} \frac{\partial \mathbf{y}_{H-1}}{\partial {\mathbf{y}_0}_i} \right) \ ,
\label{dxdy_type2}
\end{equation}
where,
\begin{equation}
\begin{aligned}
\frac{\partial \mathbf{x}_{H - 1}}{\partial {\mathbf{y}_0}_i} &= \left( \sigma'_{mc} \left(\mathbb{F}_{H-2} \right)^T \right) \circ \left(\mathbf{W}_{H-2}^{[xx]} \frac{\partial \mathbf{x}_{H-2}}{\partial {\mathbf{y}_0}_i} + \mathbf{W}_{H-2}^{[xy]} \frac{\partial \mathbf{y}_{H-2}}{\partial {\mathbf{y}_0}_i} \right) \\
    &\vdots\\
\frac{\partial \mathbf{x}_2}{\partial {\mathbf{y}_0}_i} &= \left( \sigma'_{mc} \left(\mathbb{F}_{1} \right)^T \right) \circ \left(\mathbf{W}_{1}^{[xx]} \frac{\partial \mathbf{x}_{1}}{\partial {\mathbf{y}_0}_i} + \mathbf{W}_{1}^{[xy]} \frac{\partial \mathbf{y}_{1}}{\partial {\mathbf{y}_0}_i} \right) \\
\frac{\partial \mathbf{x}_1}{\partial {\mathbf{y}_0}_i} &= \left( \sigma'_{mc} \left(\mathbb{F}_{0} \right)^T\right) \circ {\mathbf{W}_{0i}}^{[xy]} \ ,
\end{aligned}
\label{firstDerivativeeqxy_type2}
\end{equation}
and,
\begin{equation}
\begin{aligned}
\frac{\partial \mathbf{y}_{H-1}}{\partial {\mathbf{y}_0}_i} &= \left( \sigma'_{mc} \left(\mathbb{G}_{H-2} \right)^T\right) \circ \left(\mathbf{W}_{H-2}^{[yy]} \frac{\partial \mathbf{y}_{H-2}}{\partial {\mathbf{y}_0}_i} \right) \\
    &\vdots\\
\frac{\partial \mathbf{y}_{2}}{\partial {\mathbf{y}_0}_i} &= \left( \sigma'_{mc} \left(\mathbb{G}_{1} \right)^T\right) \circ \left(\mathbf{W}_{1}^{[yy]} \frac{\partial \mathbf{y}_{1}}{\partial {\mathbf{y}_0}_i} \right)\\
    \frac{\partial \mathbf{y}_1}{\partial {\mathbf{y}_0}_i} &= \left( \sigma'_{mc} \left(\mathbb{G}_{0} \right)^T\right) \circ {\mathbf{W}_{0}}_i^{[yy]} \ .
\end{aligned}
\label{firstDerivativeeqy_type2}
\end{equation}
Now we calculate the second derivatives with respect to the inputs $\mathbf{y}_0$ as:
\begin{equation}
\begin{aligned}
\frac{\partial^2 \mathbf{x}_{H}}{\partial{\mathbf{y}_0}_i \partial{\mathbf{y}_0}_j} &= \left( \sigma''_{mc} \left(\mathbb{F}_{H-1} \right)^T \right) \circ \left(\mathbf{W}_{H-1}^{[xx]} \frac{\partial \mathbf{x}_{H-1}}{\partial {\mathbf{y}_0}_i} + \mathbf{W}_{H-1}^{[xy]} \frac{\partial \mathbf{y}_{H-1}}{\partial {\mathbf{y}_0}_i} \right) \circ \left(\mathbf{W}_{H-1}^{[xx]} \frac{\partial \mathbf{x}_{H-1}}{\partial {\mathbf{y}_0}_j} + \mathbf{W}_{H-1}^{[xy]} \frac{\partial \mathbf{y}_{H-1}}{\partial {\mathbf{y}_0}_j} \right)  \\
&+ \left( \sigma'_{mc} \left(\mathbb{F}_{H -1} \right)^T\right) \circ \left(\mathbf{W}_{H -1}^{[xx]} \frac{\partial^2 \mathbf{x}_{H - 1}}{\partial{\mathbf{y}_0}_i \partial{\mathbf{y}_0}_j} + \mathbf{W}_{H -1}^{[xy]} \frac{\partial^2 \mathbf{y}_{H - 1}}{\partial{\mathbf{y}_0}_i \partial{\mathbf{y}_0}_j} \right) \\
\frac{\partial^2 \mathbf{x}_{H-1}}{\partial{\mathbf{y}_0}_i \partial{\mathbf{y}_0}_j} &= \left( \sigma''_{mc} \left(\mathbb{F}_{H-2} \right)^T \right) \circ \left(\mathbf{W}_{H-2}^{[xx]} \frac{\partial \mathbf{x}_{H-2}}{\partial {\mathbf{y}_0}_i} + \mathbf{W}_{H-2}^{[xy]} \frac{\partial \mathbf{y}_{H-2}}{\partial {\mathbf{y}_0}_i} \right) \circ \left(\mathbf{W}_{H-2}^{[xx]} \frac{\partial \mathbf{x}_{H-2}}{\partial {\mathbf{y}_0}_j} + \mathbf{W}_{H-2}^{[xy]} \frac{\partial \mathbf{y}_{H-2}}{\partial {\mathbf{y}_0}_j} \right)  \\
&+ \left( \sigma'_{mc} \left(\mathbb{F}_{H -2} \right)^T\right) \circ \left(\mathbf{W}_{H -2}^{[xx]} \frac{\partial^2 \mathbf{x}_{H - 2}}{\partial{\mathbf{y}_0}_i \partial{\mathbf{y}_0}_j} + \mathbf{W}_{H -2}^{[xy]} \frac{\partial^2 \mathbf{y}_{H - 2}}{\partial{\mathbf{y}_0}_i \partial{\mathbf{y}_0}_j} \right) \\
    &\vdots\\
\frac{\partial^2 \mathbf{x}_{2}}{\partial{\mathbf{y}_0}_i \partial{\mathbf{y}_0}_j} &= \left( \sigma''_{mc} \left(\mathbb{F}_{1} \right)^T \right) \circ \left(\mathbf{W}_{1}^{[xx]} \frac{\partial \mathbf{x}_{1}}{\partial {\mathbf{y}_0}_i} + \mathbf{W}_{1}^{[xy]} \frac{\partial \mathbf{y}_{1}}{\partial {\mathbf{y}_0}_i} \right) \circ \left(\mathbf{W}_{1}^{[xx]} \frac{\partial \mathbf{x}_{1}}{\partial {\mathbf{y}_0}_j} + \mathbf{W}_{1}^{[xy]} \frac{\partial \mathbf{y}_{1}}{\partial {\mathbf{y}_0}_j} \right)  \\
&+ \left( \sigma'_{mc} \left(\mathbb{F}_{1} \right)^T\right) \circ \left(\mathbf{W}_{1}^{[xx]} \frac{\partial^2 \mathbf{x}_{1}}{\partial{\mathbf{y}_0}_i \partial{\mathbf{y}_0}_j} + \mathbf{W}_{1}^{[xy]} \frac{\partial^2 \mathbf{y}_{1}}{\partial{\mathbf{y}_0}_i \partial{\mathbf{y}_0}_j} \right) \\
    \frac{\partial^2 \mathbf{x}_1}{\partial{\mathbf{y}_0}_i \partial{\mathbf{y}_0}_j} &= \left( \sigma'_{mc} \left(\mathbb{F}_{0} \right)^T\right) \circ {\mathbf{W}_{0i}}^{[xy]} \circ {\mathbf{W}_{0j}}^{[xy]} \ , \\
\end{aligned}
\label{secondDerivativeeqxy_type2}
\end{equation}
and,
\begin{equation}
\begin{aligned}
\frac{\partial^2 \mathbf{y}_{H-1}}{\partial{\mathbf{y}_0}_i \partial{\mathbf{y}_0}_j} &= \left( \sigma''_{mc} \left(\mathbb{G}_{H - 2} \right)^T\right) \circ \left(\mathbf{W}_{H - 2}^{[yy]} \frac{\partial \mathbf{y}_{H -2}}{\partial{\mathbf{y}_0}_i}\right) \circ \left(\mathbf{W}_{H-2}^{[yy]} \frac{\partial \mathbf{y}_{H-2}}{\partial{\mathbf{y}_0}_j} \right) \\
&+ \left( \sigma'_{mc} \left(\mathbb{G}_{H-2} \right)^T\right) \circ \left(\mathbf{W}_{H-2}^{[yy]} \frac{\partial^2 \mathbf{y}_{H-2}}{\partial{\mathbf{y}_0}_i \partial{\mathbf{y}_0}_j} \right) \\
    &\vdots\\
\frac{\partial^2 \mathbf{y}_{2}}{\partial{\mathbf{y}_0}_i \partial{\mathbf{y}_0}_j} &= \left( \sigma''_{mc} \left(\mathbb{G}_{1} \right)^T\right) \circ \left(\mathbf{W}_{1}^{[yy]} \frac{\partial \mathbf{y}_{1}}{\partial{\mathbf{y}_0}_i}\right) \circ \left(\mathbf{W}_{1}^{[yy]} \frac{\partial \mathbf{y}_{1}}{\partial{\mathbf{y}_0}_j} \right) \\
&+ \left( \sigma'_{mc} \left(\mathbb{G}_{1} \right)^T\right) \circ \left(\mathbf{W}_{1}^{[yy]} \frac{\partial^2 \mathbf{y}_{1}}{\partial{\mathbf{y}_0}_i \partial{\mathbf{y}_0}_j} \right) \\
    \frac{\partial^2 \mathbf{y}_1}{\partial{\mathbf{y}_0}_i \partial{\mathbf{y}_0}_j} &= \left( \sigma''_{mc} \left(\mathbb{G}_{0} \right)^T\right) \circ {\mathbf{W}_{0}}_i^{[yy]} \circ {\mathbf{W}_{0}}_j^{[yy]} \ .
\end{aligned}
\label{secondDerivativeeqy_type2}
\end{equation}
Similarly, we can evaluate the derivatives of the output with respect to $\mathbf{t}_0$ and $\mathbf{z}_0$.

\setcounter{equation}{0} 
\section{Model comparison for inverse design for parameters inside the training data}\label{AppC}
\begin{figure}
    \begin{subfigure}{0.5\linewidth}
        \includegraphics[scale=0.35]{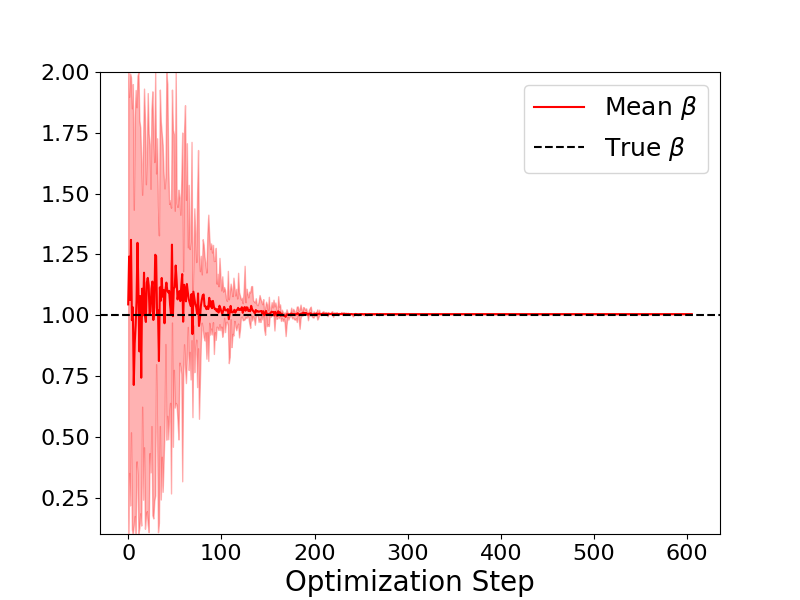}
    \caption{}
    \end{subfigure}
        \begin{subfigure}{0.5\linewidth}
        \includegraphics[scale=0.35]{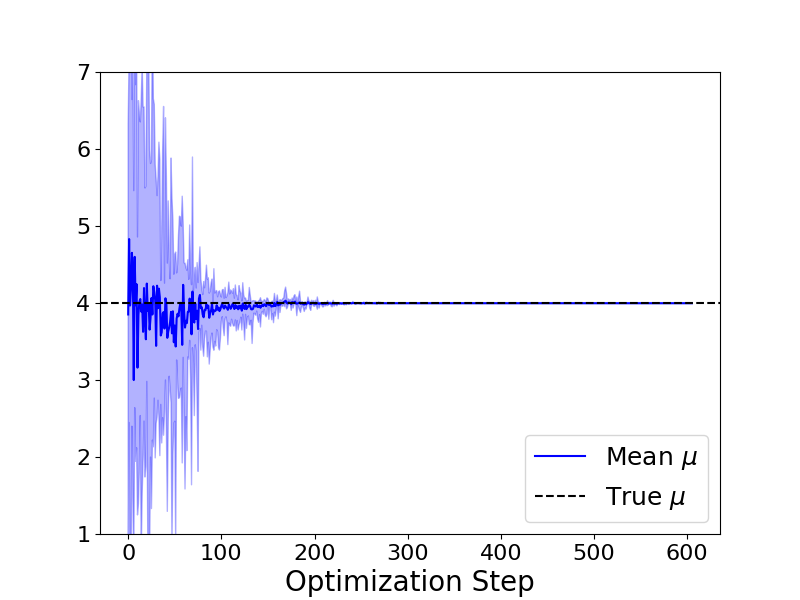}
    \caption{}
    \end{subfigure}
    \caption{The evolution for (a) $\beta$ and (b) $\mu$ for the inverse problem over 10 different initializations using CMA-ES for the parametrized potential trained using pICNN.}
    \label{inverse_pICNN_intp}
\end{figure}

\begin{figure}
    \begin{subfigure}{0.5\linewidth}
        \includegraphics[scale=0.35]{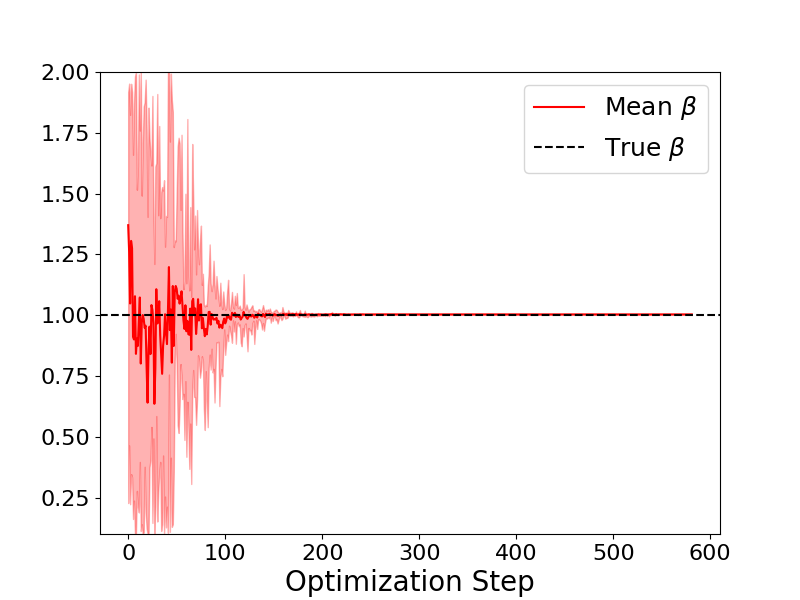}
    \caption{}
    \end{subfigure}
        \begin{subfigure}{0.5\linewidth}
        \includegraphics[scale=0.35]{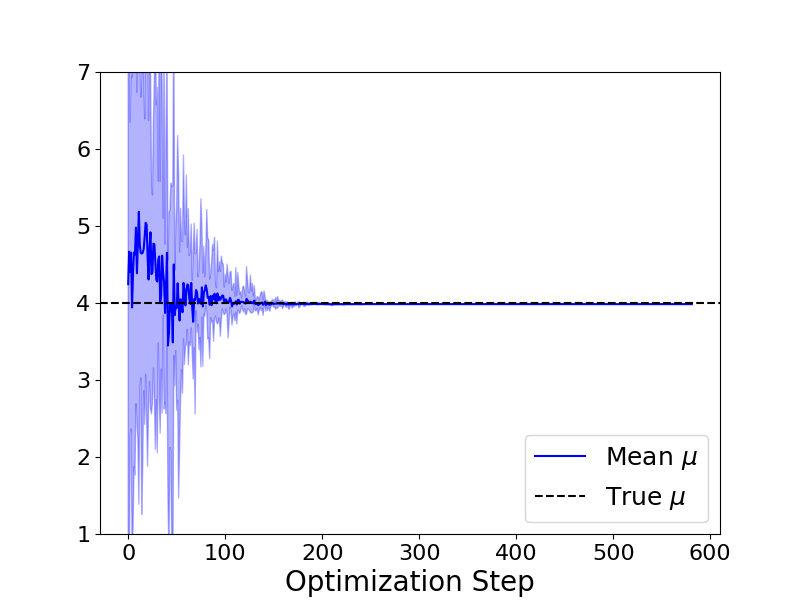}
    \caption{}
    \end{subfigure}
    \caption{The evolution for (a) $\beta$ and (b) $\mu$ for the inverse problem over 10 different initializations using CMA-ES for the parametrized potential trained using ISNN-1.}
    \label{inverse_ISNN1_intp}
\end{figure}

\begin{figure}
    \begin{subfigure}{0.5\linewidth}
        \includegraphics[scale=0.35]{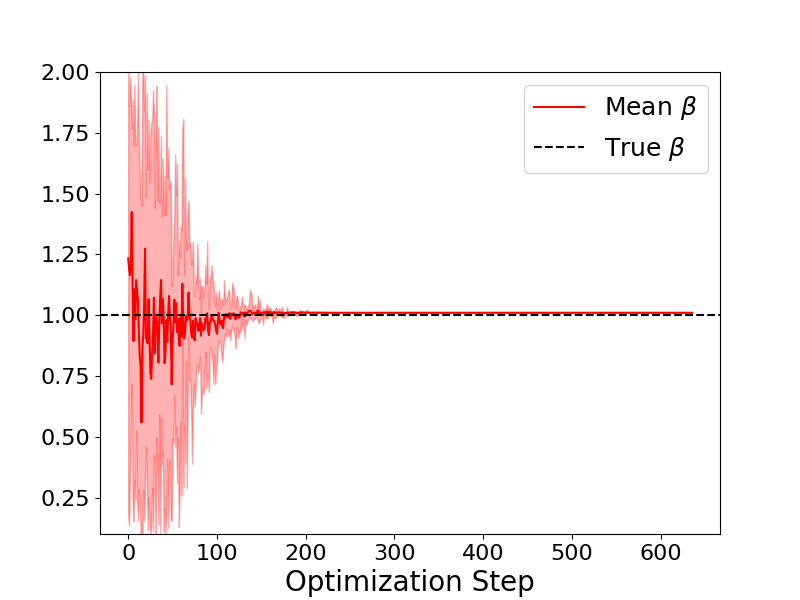}
    \caption{}
    \end{subfigure}
        \begin{subfigure}{0.5\linewidth}
        \includegraphics[scale=0.35]{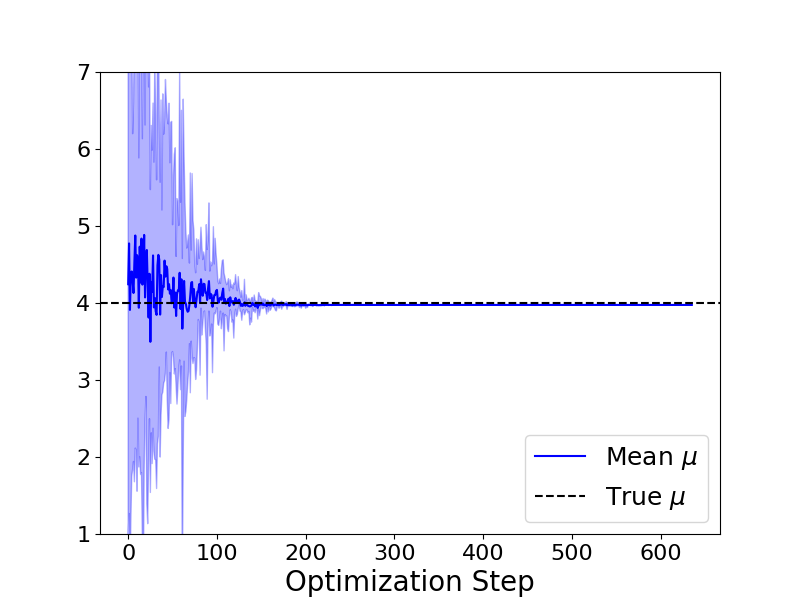}
    \caption{}
    \end{subfigure}
    \caption{The evolution for (a) $\beta$ and (b) $\mu$ for the inverse problem over 10 different initializations using CMA-ES for the parametrized potential trained using ISNN-2.}
    \label{inverse_ISNN2_intp}
\end{figure}

Here we present the inverse problem results for $\beta$ and $\mu$ values seen by the networks during training. Results for pICNN-based models are presented in Figure \ref{inverse_pICNN_intp} whereas Figures \ref{inverse_ISNN1_intp} and \ref{inverse_ISNN2_intp} present the results for ISNN-1 and ISNN-2 respectively. We can see that all the networks can invert parameters that lie within the training domain.

\clearpage
\bibliographystyle{unsrt}
\bibliography{references.bib}
\end{document}